\documentclass{article}

\usepackage{natbib}

\usepackage[preprint]{nips_2018}

\usepackage[utf8]{inputenc} 
\usepackage[T1]{fontenc}    
\usepackage{hyperref}       
\usepackage{url}            
\usepackage{booktabs}       
\usepackage{amsfonts}       
\usepackage{nicefrac}       
\usepackage{microtype}      
\usepackage{amsmath}
\usepackage{amssymb}
\usepackage{amsthm}
\usepackage{algorithm}
\usepackage{algorithmic}
\usepackage{graphicx}
\usepackage{subcaption}
\usepackage{wrapfig}
\usepackage{placeins}

\title{Variational Option Discovery Algorithms}

\author{
  Joshua Achiam \\
  UC Berkeley \& OpenAI\\ 
  \And
  Harrison Edwards\\
  OpenAI \\
  \And
  Dario Amodei \\
  OpenAI
  \And
  Pieter Abbeel \\
  UC Berkeley
}

\newcommand{\underE}[2]{\underset{\begin{subarray}{c}#1 \end{subarray}}{\E}\left[ #2 \right]}

\begin{document}

%
%



\newcommand{\avet}{{\mathbf  a}}
\newcommand{\bvet}{{\mathbf  b}}
\newcommand{\cvet}{{\mathbf  c}}
\newcommand{\dvet}{{\mathbf  d}}
\newcommand{\evet}{{\mathbf  e}}
\newcommand{\fvet}{{\mathbf  f}}
\newcommand{\gvet}{{\mathbf  g}}
\newcommand{\hvet}{{\mathbf  h}}
\newcommand{\ivet}{{\mathbf  i}}
\newcommand{\jvet}{{\mathbf  j}}
\newcommand{\kvet}{{\mathbf  k}}
\newcommand{\lvet}{{\mathbf  l}}
\newcommand{\mvet}{{\mathbf  m}}
\newcommand{\nvet}{{\mathbf  n}}
\newcommand{\ovet}{{\mathbf  o}}
\newcommand{\pvet}{{\mathbf  p}}
\newcommand{\qvet}{{\mathbf  q}}
\newcommand{\rvet}{{\mathbf  r}}
\newcommand{\svet}{{\mathbf  s}}
\newcommand{\tvet}{{\mathbf  t}}
\newcommand{\uvet}{{\mathbf  u}}
\newcommand{\vvet}{{\mathbf  v}}
\newcommand{\xvet}{{\mathbf  x}}
\newcommand{\yvet}{{\mathbf  y}}
\newcommand{\zvet}{{\mathbf  z}}
\newcommand{\wvet}{{\mathbf  w}}

\newcommand{\Avet}{{\mathbf  A}}
\newcommand{\Bvet}{{\mathbf  B}}
\newcommand{\Cvet}{{\mathbf  C}}
\newcommand{\Dvet}{{\mathbf  D}}
\newcommand{\Evet}{{\mathbf  E}}
\newcommand{\Fvet}{{\mathbf  F}}
\newcommand{\Gvet}{{\mathbf  G}}
\newcommand{\Hvet}{{\mathbf  H}}
\newcommand{\Ivet}{{\mathbf  I}}
\newcommand{\Jvet}{{\mathbf  J}}
\newcommand{\Kvet}{{\mathbf  K}}
\newcommand{\Lvet}{{\mathbf  L}}
\newcommand{\Mvet}{{\mathbf  M}}
\newcommand{\Nvet}{{\mathbf  N}}
\newcommand{\Ovet}{{\mathbf  O}}
\newcommand{\Pvet}{{\mathbf  P}}
\newcommand{\Qvet}{{\mathbf  Q}}
\newcommand{\Rvet}{{\mathbf  R}}
\newcommand{\Svet}{{\mathbf  S}}
\newcommand{\Tvet}{{\mathbf  T}}
\newcommand{\Uvet}{{\mathbf  U}}
\newcommand{\Xvet}{{\mathbf  X}}
\newcommand{\Yvet}{{\mathbf  Y}}
\newcommand{\Vvet}{{\mathbf  V}}
\newcommand{\Wvet}{{\mathbf  W}}
\newcommand{\Zvet}{{\mathbf  Z}}

\newcommand{\Deltavet}{\mathbf  \Delta}
\newcommand{\Lambdavet}{{\mathbf  \Lambda}}
\newcommand{\Sigmavet}{\mathbf  \Sigma}
\newcommand{\Thetavet}{{\mathbf  \Theta}}

\newcommand{\s}{ {\sigma} }

\newcommand{\e}{{\mathrm e}}
\newcommand{\jm}{{\mathrm j}}
\newcommand{\E}{{\mathrm E}}
\newcommand{\Ex}{{\mathbb E}}
\renewcommand{\d}{{\mathrm d}}
\newcommand{\dt}{{\mathrm d}t}
\newcommand{\X}{ {\mathcal X} }
\newcommand{\Y}{ {\mathcal Y} }
\newcommand{\Z}{ {\mathcal Z} }

\newcommand{\calA}{{\mathcal A}}
\newcommand{\calB}{{\mathcal B}}
\newcommand{\calC}{{\mathcal C}}
\newcommand{\calD}{{\mathcal D}}
\newcommand{\calE}{{\mathcal E}}
\newcommand{\calF}{{\mathcal F}}
\newcommand{\calG}{{\mathcal G}}
\newcommand{\calH}{{\mathcal H}}
\newcommand{\calI}{{\mathcal I}}
\newcommand{\calJ}{{\mathcal J}}
\newcommand{\calK}{{\mathcal K}}
\newcommand{\calL}{{\mathcal L}}
\newcommand{\calM}{{\mathcal M}}
\newcommand{\calN}{{\mathcal N}}
\newcommand{\calO}{{\mathcal O}}
\newcommand{\calP}{{\mathcal P}}
\newcommand{\calQ}{{\mathcal Q}}
\newcommand{\calR}{{\mathcal R}}
\newcommand{\calS}{{\mathcal S}}
\newcommand{\calT}{{\mathcal T}}
\newcommand{\calU}{{\mathcal U}}
\newcommand{\calV}{{\mathcal V}}
\newcommand{\calX}{{\mathcal X}}
\newcommand{\calY}{{\mathcal Y}}
\newcommand{\calW}{{\mathcal W}}
\newcommand{\calZ}{{\mathcal Z}}
\newcommand{\qtil}{{\tilde{q}}}
\newcommand{\td}{{\tilde{\delta}}}

\newcommand{\vect}[1]{ {\mbox{\rm vec}(#1)} }


\newcommand{\Atil}{\tilde{A}}
\newcommand{\Zhat}{\hat{Z}}
\newcommand{\Hbar}{\bar{H}}
\newcommand{\Dhat}{\hat{D}}
\newcommand{\dhat}{\hat{d}}

\newcommand{\rhat}{\hat{r}}
\newcommand{\xhat}{\hat{x}}
\newcommand{\yhat}{\hat{y}}
\newcommand{\zhat}{\hat{z}}
\newcommand{\xbar}{\bar{x}}
\newcommand{\ubar}{\bar{u}}
\newcommand{\ybar}{\bar{y}}
\newcommand{\zbar}{\bar{z}}
\newcommand{\pdot}{\dot{p}}
\newcommand{\pddot}{\ddot{p}}
\newcommand{\pbar}{\bar{p}}
\newcommand{\qdot}{\dot{q}}
\newcommand{\qddot}{\ddot{q}}
\newcommand{\qbar}{\bar{q}}
\newcommand{\xdot}{\dot{x}}
\newcommand{\ydot}{\dot{y}}
\newcommand{\zdot}{\dot{z}}
\newcommand{\yddot}{\ddot{y}}
\newcommand{\thdot}{\dot{\theta}}
\newcommand{\thddot}{\ddot{\theta}}
\newcommand{\util}{{\tilde{u}}}
\newcommand{\xtil}{{\tilde{x}}}
\newcommand{\ytil}{{\tilde{y}}}
\newcommand{\lam}{\lambda}
\newcommand{\lamax}{\lambda\ped{max}}
\newcommand{\lamin}{\lambda\ped{min}}
\newcommand{\adj}{ {\mbox{\rm adj}\;} }
\newcommand{\sign}{\mbox {\rm sgn}}
\newcommand{\spn}{\mbox {\rm span}}
\newcommand{\barJ}{\bar{J}}
\newcommand{\dom}{\mathop {\mathrm {dom}}}
\newcommand{\card}{\mathop{\mathrm{card}}}
\newcommand{\subt}{\mathop{\mathrm{s.t.}}}

\newcommand{\epi}{\mathop{\mathrm{epi}}}
\newcommand{\env}{\mathop{\mathrm{env}}}
\newcommand{\chull}{\mathop{\mathrm{co}}}
\newcommand{\graph}{\mathop{\mathrm{graph}}}
\newcommand{\prox}[1]{\mathop{\mathrm{prox}_{#1}}}
\newcommand{\sthr}[1]{\mathop{\mathrm{sthr}_{#1}}}

\def\hardsection{$\spadesuit\;$}

\newcommand{\Real}[1]{ { {\mathbb R}^{#1} } }
\newcommand{\Realp}[1]{ { {\mathbb R}_{+}^{#1} } }
\newcommand{\Realpp}[1]{ { {\mathbb R}_{++}^{#1} } }
\newcommand{\Complex}[1]{ { {\mathbb C}^{#1} } }
\newcommand{\Imag}[1]{ { {\mathbb I}^{#1} } }
\newcommand{\Field}[1]{ {\mathbb F}^{#1} }
\newcommand{\F}{ {\mathbb F}}
\newcommand{\Orth}[1]{ { {\calG_{\calO}^{#1}} } }
\newcommand{\Unit}[1]{ { {\calG_{\calU}^{#1}} } }
\newcommand{\Sym}[1]{ { {\mathbb S}^{#1} } }
\newcommand{\Symp}[1]{ { {\mathbb S}_{+}^{#1} } }
\newcommand{\Sympp}[1]{ { {\mathbb S}_{++}^{#1} } }
\newcommand{\Herm}[1]{ { {\mathbb H}^{#1} } }
\newcommand{\Skew}[1]{ { {\mathbb S\mathbb K}^{#1} } }
\newcommand{\Skherm}[1]{ { {\mathbb H\mathbb K}^{#1} } }
\newcommand{\Rman}[1]{ { {\mathcal R}^{#1} } } 
\newcommand{\Cman}[1]{ { {\mathcal C}^{#1} } }
\newcommand{\Hinf}[1]{ {  {\mathcal H}_\infty^{#1} } }
\newcommand{\RHinf}[1]{ { {\mathcal RH}_\infty^{#1} } }
\newcommand{\Htwo}[1]{ {  {\mathcal H}_2^{#1} } }
\newcommand{\RHtwo}[1]{ { {\mathcal RH}_2^{#1} } }

\newcommand{\dist}[1]{{\mathrm{dist}}{\left( #1 \right)}}
\newcommand{\diff}[2]{ \frac{\d {#1}}{\d {#2}}  }
\newcommand{\diffp}[2]{ \frac{\partial {#1}}{\partial {#2}}  }
\newcommand{\diffqd}[2]{ \frac{\d^2 {#1}}{\d {#2}^2}  }
\newcommand{\diffq}[2]{ \frac{\d^2 {#1}}{\d {#2}}  }
\newcommand{\diffqq}[3]{ \frac{\d^2 {#1}}{ \d {#2} \d {#3}  }}
\newcommand{\diffpq}[2]{ \frac{\partial^2 {#1}}{\partial {#2}^2}  }
\newcommand{\difftq}[3]{ \frac{\partial^2 {#1}}{\partial {#2}\partial {#3}}  }
\newcommand{\diffi}[3]{ \frac{\d^{#3} {#1}}{\d {#2}^{#3}}  }
\newcommand{\diffpi}[3]{ \frac{\partial^{#3} {#1}}{\partial {#2}^{#3}}  }
\newcommand{\binomial}[2]{\scriptsize{\left(\!\! \ba{c} #1 \\ #2 \ea \!\! \right)} }
\newcommand{\comb}[2]{{\left(\!\!\! \ba{c} #1 \\ #2 \ea \!\!\! \right)} }

\newcommand{\simax}{{\sigma_{\mathrm{max}}}}
\newcommand{\simin}{{\sigma_{\mathrm{min}}}}
\newcommand{\prob}{{\mbox{\rm Prob}}}
\newcommand{\var}{{\mbox{\rm var}}}
\newcommand{\sint}{{\mbox{\rm int}\,}} 
\newcommand{\relint}{{\mbox{\rm relint}\,}} 
\newcommand{\ns}{{\mbox{\tt ns}}}

\newcommand{\rank}{\mathop{\mathrm{rank}}\nolimits}
\newcommand{\range}{\mathop{\mathcal{R}}\nolimits}
\newcommand{\nulsp}{\mathop{\mathcal{N}}\nolimits}
\newcommand{\diagop}{\mathop{\mathrm{diag}}\nolimits}
\newcommand{\Var}{\mathop{\mathrm{var}}\nolimits}
\newcommand{\tr}{\mathop{\mathrm{trace}}\nolimits}
\newcommand{\sinc}{\mathop{\mathrm{sinc}}\nolimits}

\newcommand{\pre}[1]{ { {\mathop{\mathrm{Re}}}  \left({#1}\right)} }
\newcommand{\pim}[1]{ { {\mathop{\mathrm{Im}}}  ({#1})} }
\newcommand{\rp}{ ^{\Real{}} }
\newcommand{\ip}{ ^{\Imag{}} }

\newcommand{\one}{{\mathbf  1}}
\newcommand{\dss}{\displaystyle}
\newcommand{\inv}{^{-1}}
\newcommand{\pinv}{^{\dagger}}
\newcommand{\diag}[1]{\mathrm{diag}\left({#1}\right)}
\newcommand{\blockdiag}[1]{\mbox{\rm bdiag}\left({#1}\right)}
\newcommand{\tran}{^{\top}}
\newcommand{\inner}[1]{\langle {#1} \rangle}
\newcommand{\ped}[1]{_{\mathrm{#1}}}
\newcommand{\ap}[1]{^{\mathrm{#1}}}

\newcommand{\blu}[1]{\textcolor{blue}{#1}}
\newcommand{\red}[1]{\textcolor{red}{#1}}
\newcommand{\green}[1]{\textcolor{green}{#1}}
\newcommand{\cyan}[1]{\textcolor{cyan}{#1}}
\newcommand{\comment}[1]{\vspace{.1cm} \blu{#1} \vspace{.1cm}}

\newcommand{\beq}{\begin{equation}}
\newcommand{\eeq}{\end{equation}}
\newcommand{\bea}{\begin{eqnarray}}
\newcommand{\eea}{\end{eqnarray}}
\newcommand{\beas}{\begin{eqnarray*}}
\newcommand{\eeas}{\end{eqnarray*}}
\newcommand{\ba}{\begin{array}}
\newcommand{\ea}{\end{array}}
\newcommand{\bit}{\begin{itemize}}
\newcommand{\eit}{\end{itemize}}
\newcommand{\ben}{\begin{enumerate}}
\newcommand{\een}{\end{enumerate}}
\newcommand{\bde}{\begin{description}}
\newcommand{\ede}{\end{description}}
\newcommand{\bsp}{\begin{split}}
\newcommand{\esp}{\end{split}}


%
%

\def\nocolon{}

\newcommand{\monthyear}{%
  \ifcase\month\or January\or February\or March\or April\or May\or June\or
  July\or August\or September\or October\or November\or
  December\fi\space\number\year
}

\newcommand{\openepigraph}[2]{%
  \begin{fullwidth}
  \sffamily\large
  \begin{doublespace}
  \noindent\allcaps{#1}\\
  \noindent\allcaps{#2}
  \end{doublespace}
  \end{fullwidth}
}

\newcommand{\blankpage}{\newpage\hbox{}\thispagestyle{empty}\newpage}

\maketitle

\begin{abstract}
We explore methods for option discovery based on variational inference and make two algorithmic contributions. First: we highlight a tight connection between variational option discovery methods and variational autoencoders, and introduce Variational Autoencoding Learning of Options by Reinforcement (VALOR), a new method derived from the connection. In VALOR, the policy encodes contexts from a noise distribution into trajectories, and the decoder recovers the contexts from the complete trajectories. Second: we propose a curriculum learning approach where the number of contexts seen by the agent increases whenever the agent's performance is strong enough (as measured by the decoder) on the current set of contexts. We show that this simple trick stabilizes training for VALOR and prior variational option discovery methods, allowing a single agent to learn many more modes of behavior than it could with a fixed context distribution. Finally, we investigate other topics related to variational option discovery, including fundamental limitations of the general approach and the applicability of learned options to downstream tasks. 
\end{abstract}

\section{Introduction}

Humans are innately driven to experiment with new ways of interacting with their environments. This can accelerate the process of discovering skills for downstream tasks and can also be viewed as a primary objective in its own right. This drive serves as an inspiration for reward-free option discovery in reinforcement learning (based on the options framework of \cite{Sutton1999, Precup2000}), where an agent tries to learn skills by interacting with its environment without trying to maximize cumulative reward for a particular task.

In this work, we explore variational option discovery, the space of methods for option discovery based on variational inference. We highlight a tight connection between prior work on variational option discovery and variational autoencoders (\citet{Kingma2013}), and derive a new method based on the connection. In our analogy, a policy acts as an encoder, translating contexts from a noise distribution into trajectories; a decoder attempts to recover the contexts from the trajectories, and rewards the policies for making contexts easy to distinguish. Contexts are random vectors which have no intrinsic meaning prior to training, but they become associated with trajectories as a result of training; each context vector thus corresponds to a distinct option. Therefore this approach learns a set of options which are as diverse as possible, in the sense of being as easy to distinguish from each other as possible. We show that Variational Intrinsic Control (VIC) (\citet{Gregor2016}) and the recently-proposed Diversity is All You Need (DIAYN) (\citet{Eysenbach2018}) are specific instances of this template which decode from states instead of complete trajectories.

We make two main algorithmic contributions:
\begin{enumerate}
\item We introduce Variational Autoencoding Learning of Options by Reinforcement (VALOR), a new method which decodes from trajectories.
The idea is to encourage learning dynamical modes instead of goal-attaining modes, e.g. `move in a circle' instead of `go to X'.
\item We propose a curriculum learning approach where the number of contexts seen by the agent increases whenever the agent's performance is strong enough (as measured by the decoder) on the current set of contexts. 
\end{enumerate}

We perform a comparison analysis of VALOR, VIC, and DIAYN with and without the curriculum trick, evaluating them in various robotics environments (point mass, cheetah, swimmer, ant).\footnote{Videos of learned behaviors will be made available at \url{varoptdisc.github.io}.} We show that, to the extent that our metrics can measure, all three of them perform similarly, except that VALOR can attain qualitatively different behavior because of its trajectory-centric approach, and DIAYN learns more quickly because of its denser reward signal. We show that our curriculum trick stabilizes and speeds up learning for all three methods, and can allow a single agent to learn up to hundreds of modes. Beyond our core comparison, we also explore applications of variational option discovery in two interesting spotlight environments: a simulated robot hand and a simulated humanoid. Variational option discovery finds naturalistic finger-flexing behaviors in the hand environment, but performs poorly on the humanoid, in the sense that it does not discover natural crawling or walking gaits. We consider this evidence that pure information-theoretic objectives can do a poor job of capturing human priors on useful behavior in complex environments. Lastly, we try a proof-of-concept for applicability to downstream tasks in a variant of ant-maze by using a (particularly good) pretrained VALOR policy as the lower level of a hierarchy. In this experiment, we find that the VALOR policy is more useful than a random network as a lower level, and equivalently as useful as learning a lower level from scratch in the environment.

\section{Related Work}

\textbf{Option Discovery}: Substantial prior work exists on option discovery (\cite{Sutton1999, Precup2000}); here we will restrict our attention to relevant recent work in the deep RL setting. \citet{Bacon2017} and \citet{Fox2017} derive policy gradient methods for learning options: \citet{Bacon2017} learn options concurrently with solving a particular task, while \citet{Fox2017} learn options from demonstrations to accelerate specific-task learning. \citet{Vezhnevets2017} propose an architecture and training algorithm which can be interpreted as implicitly learning options. \cite{Thomas2017} find options as controllable factors in the environment. \cite{Machado2017}, \cite{Machado2017a}, and \cite{Liu2017} learn \textit{eigenoptions}, options derived from the graph Laplacian associated with the MDP. Several approaches for option discovery are primarily information-theoretic: \citet{Gregor2016}, \citet{Eysenbach2018}, and \citet{Florensa2017} train policies to maximize mutual information between options and states or quantities derived from states; by contrast, we maximize information between options and whole trajectories. \citet{Hausman2018} learn skill embeddings by optimizing a variational bound on the entropy of the policy; the final objective function is closely connected with that of \citet{Florensa2017}. 

\textbf{Universal Policies}: Variational option discovery algorithms learn universal policies (goal- or instruction- conditioned policies), like universal value function approximators (\cite{Schaul2015}) and hindsight experience replay (\cite{Andrychowicz2017}). However, these other approaches require extrinsic reward signals and a hand-crafted instruction space. By contrast, variational option discovery is unsupervised and finds its own instruction space. 

\textbf{Intrinsic Motivation}: Many recent works have incorporated intrinsic motivation (especially curiosity) into deep RL agents (\cite{Stadie2015, Houthooft2016, Bellemare2016, Achiam2017, Fu2017, Pathak2017, Ostrovski2017, Edwards2018}). However, none of these approaches were combined with learning universal policies, and so suffer from a problem of knowledge fade: when states cease to be interesting to the intrinsic reward signal (usually when they are no longer novel), unless they coincide with extrinsic rewards or are on a direct path to the next-most novel state, the agent will forget how to visit them. 

\textbf{Variational Autoencoders}: Variational autoencoders (VAEs) (\cite{Kingma2013}) learn a probabilistic encoder $q_{\phi} (z|x)$ and decoder $p_{\theta}(x|z)$ which map between data $x$ and latent variables $z$ by optimizing the evidence lower bound (ELBO) on the marginal distribution $p_{\theta}(x)$, assuming a prior $p(z)$ over latent variables. \cite{Higgins2017} extended the VAE approach by including a parameter $\beta$ to control the capacity of $z$ and improve the ability of VAEs to learn disentangled representations of high-dimensional data. The $\beta$-VAE optimization problem is
\begin{equation}
\max_{\phi, \theta} \underE{x \sim \calD}{ \underE{z \sim q_{\phi}(\cdot|x)}{\log p_{\theta}(x|z)} - \beta D_{KL}\left( q_{\phi}(z|x) || p(z)\right)}, \label{betavae} 
\end{equation}
and when $\beta = 1$, it reduces to the standard VAE of \citet{Kingma2013}.

\textbf{Novelty Search}: Option discovery algorithms based on the diversity of learned behaviors can be viewed as similar in spirit to novelty search (\cite{Lehman2012}), an evolutionary algorithm which finds behaviors which are diverse with respect to a characterization function which is usually pre-designed but sometimes learned (as in \cite{Meyerson2016}).

\section{Variational Option Discovery Algorithms}

Our aim is to learn a policy $\pi$ where action distributions are conditioned on both the current state $s_t$ and a \textit{context} $c$ which is sampled at the start of an episode and kept fixed throughout. The context should uniquely specify a particular mode of behavior (also called a skill). But instead of using reward functions to ground contexts to trajectories, we want the meaning of a context to be arbitrarily assigned (`discovered') during training. 

We formulate a learning approach as follows. A context $c$ is sampled from a noise distribution $G$, and then encoded into a trajectory $\tau = (s_0, a_0, ..., s_T)$ by a policy $\pi(\cdot|s_t, c)$; afterwards $c$ is decoded from $\tau$ with a probabilistic decoder $D$. If the trajectory $\tau$ is unique to $c$, the decoder will place a high probability on $c$, and the policy should be correspondingly reinforced. Supervised learning can be applied to the decoder (because for each $\tau$, we know the ground truth $c$). To encourage exploration, we include an entropy regularization term with coefficient $\beta$. The full optimization problem is thus
\begin{equation}
\max_{\pi, D} \; \underE{c \sim G}{\underE{\tau \sim \pi, c}{\log P_D (c | \tau)} + \beta \calH(\pi| c) }, \label{vaerl}
\end{equation}
where $P_D$ is the distribution over contexts from the decoder, and the entropy term is $\calH(\pi| c) \doteq \E_{\tau \sim \pi,c}\left[\sum_{t} H(\pi(\cdot|s_t, c))\right]$.
We give a generic template for option discovery based on Eq. \ref{vaerl} as Algorithm \ref{alg0}. Observe that the objective in Eq. \ref{vaerl} has a one-to-one correspondence with the $\beta$-VAE objective in Eq. \ref{betavae}: the context $c$ maps to the data $x$, the trajectory $\tau$ maps to the latent representation $z$, the policy $\pi$ and the MDP together form the encoder $q_{\phi}$, the decoder $D$ maps to the decoder $p_{\theta}$, and the entropy regularization $\calH(\pi|c)$ maps to the KL-divergence of the encoder distribution from a prior where trajectories are generated by a uniform random policy (proof in Appendix A). Based on this connection, we call algorithms for solving Eq. \ref{vaerl} variational option discovery methods.

\begin{algorithm}
   \caption{Template for Variational Option Discovery with Autoencoding Objective}
   \label{alg0}
\begin{algorithmic}
	\STATE Generate initial policy $\pi_{\theta_0}$, decoder $D_{\phi_0}$

	 \FOR{$k = 0,1,2,...$} 
	 \STATE Sample context-trajectory pairs $\calD = \{(c^i, \tau^i)\}_{i=1, ..., N}$, by first sampling a context $c \sim G$ and then rolling out a trajectory in the environment, $\tau \sim \pi_{\theta_k}(\cdot|\cdot, c)$.
	 \STATE Update policy with any reinforcement learning algorithm to maximize Eq. \ref{vaerl}, using batch $\calD$
	 \STATE Update decoder by supervised learning to maximize $\underE{}{\log P_D (c|\tau)}$, using batch $\calD$
	\ENDFOR
\end{algorithmic}
\end{algorithm}

\subsection{Connections to Prior Work}

\textbf{Variational Intrinsic Control}: Variational Intrinsic Control\footnote{Specifically, the algorithm presented as `Intrinsic Control with Explicit Options' in \citet{Gregor2016}.} (VIC) (\citet{Gregor2016}) is an option discovery technique based on optimizing a variational lower bound on the mutual information between the context and the final state in a trajectory, conditioned on the initial state. \citet{Gregor2016} give the optimization problem as
\begin{equation}
\max_{G, \pi, D} \; \underE{s_0 \sim \mu}{\underE{c \sim G(\cdot|s_0) \\ \tau \sim \pi, c}{\log P_D (c | s_0, s_T)} + H(G(\cdot|s_0))}, \label{VIC}
\end{equation}
where $\mu$ is the starting state distribution for the MDP. This differs from Eq. \ref{vaerl} in several ways: the context distribution $G$ can be optimized, $G$ depends on the initial state $s_0$, $G$ is entropy-regularized, entropy regularization for the policy $\pi$ is omitted, and the decoder only looks at the first and last state of the trajectory instead of the entire thing. However, they also propose to keep $G$ fixed and state-independent, and do this in their experiments; additionally, their experiments use decoders which are conditioned on the final state only. This reduces Eq. \ref{VIC} to Eq. \ref{vaerl} with $\beta = 0$ and $\log P_D (c| \tau) = \log P_D (c| s_T)$. We treat this as the canonical form of VIC and implement it this way for our comparison study.

\textbf{Diversity is All You Need}: Diversity is All You Need (DIAYN) (\citet{Eysenbach2018}) performs option discovery by optimizing a variational lower bound for an objective function designed to maximize mutual information between context and \textit{every} state in a trajectory, while minimizing mutual information between actions and contexts conditioned on states, and maximizing entropy of the mixture policy over contexts. The exact optimization problem is
\begin{equation}
\max_{\pi, D} \; \underE{c \sim G}{\underE{\tau \sim \pi, c}{ \sum_{t=0}^T \left(\log P_D (c|s_t) - \log G(c)\right)} + \beta \calH(\pi| c) }. \label{diayn}
\end{equation} 
In DIAYN, $G$ is kept fixed (as in canonical VIC), so the term $\log G(c)$ is constant and may be removed from the optimization problem. Thus Eq. \ref{diayn} is a special case of Eq. \ref{vaerl} with $\log P_D (c|\tau) = \sum_{t=0}^T \log P_D (c|s_t)$. 

\subsection{VALOR}

\begin{wrapfigure}{R}{0.4\textwidth}
  \begin{center}
    \includegraphics[width=0.4\textwidth]{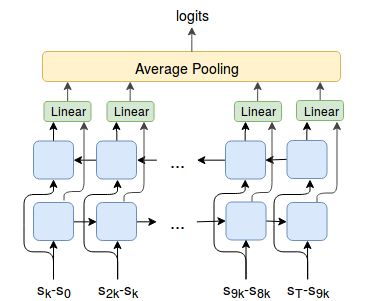}
  \end{center}
  \caption{Bidirectional LSTM architecture for VALOR decoder. Blue blocks are LSTM cells.}
  \label{valor_decoder}
\end{wrapfigure}

In this section, we propose Variational Autoencoding Learning of Options by Reinforcement (VALOR), a variational option discovery method which directly optimizes Eq. \ref{vaerl} with two key decisions about the decoder:
\begin{itemize}
\item The decoder never sees actions. Our conception of `interesting' behaviors requires that the agent attempt to interact with the environment to achieve some change in state. If the decoder was permitted to see raw actions, the agent could signal the context directly through its actions and ignore the environment. Limiting the decoder in this way forces the agent to manipulate the environment to communicate with the decoder. 
\item Unlike in DIAYN, the decoder does \textit{not} decompose as a sum of per-timestep computations. That is, $\log P_{D} (c|\tau) \neq \sum_{t=0}^T f(s_t, c)$. We choose against this decomposition because it could limit the ability of the decoder to correctly distinguish between behaviors which share some states, or behaviors which share all states but reach them in different orders. 
\end{itemize}

We implement VALOR with a recurrent architecture for the decoder (Fig. \ref{valor_decoder}), using a bidirectional LSTM to make sure that both the beginning and end of a trajectory are equally important. We only use $N=11$ equally spaced observations from the trajectory as inputs, for two reasons: 1) computational efficiency, and 2) to encode a heuristic that we are only interested in low-frequency behaviors (as opposed to information-dense high-frequency jitters). Lastly, taking inspiration from \citet{Vezhnevets2017}, we only decode from the $k$-step \textit{transitions} (deltas) in state space between the $N$ observations. Intuitively, this corresponds to a prior that agents should move, as any two modes where the agent stands still in different poses will be indistinguishable to the decoder (because the deltas will be identically zero). We do not decode from transitions in VIC or DIAYN, although we note it would be possible and might be interesting future work. 

\subsection{Curriculum Approach}

The standard approach for context distributions, used in VIC and DIAYN, is to have $K$ discrete contexts with a uniform distribution: $c \sim \text{Uniform}(K)$. In our experiments, we found that this worked poorly for large $K$ across all three algorithms we compared. Even with very large batches (to ensure that each context was sampled often enough to get a low-variance contribution to the gradient), training was challenging. We found a simple trick to resolve this issue: start training with small $K$ (where learning is easy), and gradually increase it over time as the decoder gets stronger. Whenever $\underE{}{\log P_D (c|\tau)}$ is high enough (we pick a fairly arbitrary threshold of $P_D (c|\tau) \approx 0.86$), we increase $K$ according to
\begin{equation}
K \leftarrow \min\left( \text{int}\left(1.5 \times K + 1\right),  K_{max}\right),
\label{curriculum}
\end{equation}
where $K_{max}$ is a hyperparameter. As our experiments show, this curriculum leads to faster and more stable convergence.

\section{Experimental Setup} 

In our experiments, we try to answer the following questions:
\begin{itemize}
\item What are best practices for training agents with variational option discovery algorithms (VALOR, VIC, DIAYN)? Does the curriculum learning approach help?
\item What are the qualitative results from running variational option discovery algorithms? Are the learned behaviors recognizably distinct to a human? Are there substantial differences between algorithms?
\item Are the learned behaviors useful for downstream control tasks?
\end{itemize}

\textbf{Test environments}: Our core comparison experiments is on a slate of locomotion environments: a custom 2D point agent, the HalfCheetah and Swimmer robots from the OpenAI Gym \citep{Brockman2016}, and a customized version of Ant from Gym where contact forces are omitted from the observations. We also tried running variational option discovery on two other interesting simulated robots: a dextrous hand (with $\calS \in \Real{48}$ and $\calA \in \Real{20}$, based on \cite{Plappert}), and a new complex humanoid environment we call `toddler' (with $\calS \in \Real{335}$ and $\calA \in \Real{35}$). Lastly, we investigated applicability to downstream tasks in a modified version of Ant-Maze (\cite{Frans2018}).

\textbf{Implementation}: We implement VALOR, VIC, and DIAYN with vanilla policy gradient as the RL algorithm (described in Appendix \ref{implementation}). We note that VIC and DIAYN were originally implemented with different RL algorithms: \citet{Gregor2016} implemented VIC with tabular Q learning (\cite{Watkins1992}), and \citet{Eysenbach2018} implemented DIAYN with soft actor-critic (\cite{Haarnoja}). Also unlike prior work, we use recurrent neural network policy architectures. Because there is not a final objective function to measure whether an algorithm has achieved qualitative diversity of behaviors, our hyperparameters are based on what resulted in stable training, and kept constant across algorithms. Because the design space for these algorithms is very large and evaluation is to some degree subjective, we caution that our results should not necessarily be viewed as definitive.

\textbf{Training techniques}: We investigated two specific techniques for training: curriculum generation via Eq. \ref{curriculum}, and context embeddings. On context embeddings: a natural approach for providing the integer context as input to a neural network policy is to convert the context to a one-hot vector and concatenate it with the state, as in \citet{Eysenbach2018}. Instead, we consider whether training is improved by allowing the agent to learn its own embedding vector for each context.

\section{Results}

\begin{figure}
\centering
\begin{subfigure}{.24\textwidth}
  \includegraphics[height=2.4cm]{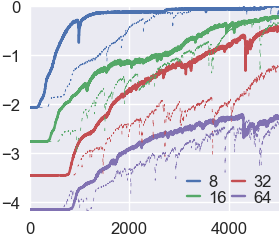}
  \caption{Uniform, for various $K$}
  \label{bp-b}
\end{subfigure}%
\begin{subfigure}{.24\textwidth}
  \includegraphics[height=2.4cm]{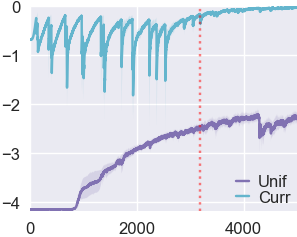}  
  \caption{Uniform vs Curriculum}
  \label{bp-c}
\end{subfigure}%
\begin{subfigure}{.24\textwidth}
  \includegraphics[height=2.4cm]{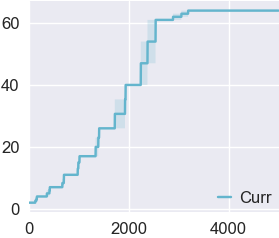}  
  \caption{Curriculum, current $K$}
  \label{bp-d}
\end{subfigure}

\caption[]{Studying optimization techniques with VALOR in HalfCheetah, showing performance---in (a) and (b), $\E[\log P_D(c|\tau)]$; in (c), the value of $K$ throughout the curriculum---vs training iteration. (a) compares learning curves with and without context embeddings (solid vs dotted, resp.), for $K\in\{8,16,32,64\}$, with uniform context distributions. (b) compares curriculum (with $K_{max}=64$) to uniform (with $K=64$) context distributions, using embeddings for both. The dips for the curriculum curve indicate when $K$ changes via Eq. \ref{curriculum}; values of $K$ are shown in (c). The dashed red line shows when $K=K_{max}$ for the curriculum; after it, the curves for Uniform and Curriculum can be fairly compared. All curves are averaged over three random seeds.}
\label{best_practices}
\end{figure}

\textbf{Exploring Optimization Techniques}: We present partial findings for our investigation of training techniques in Fig. \ref{best_practices} (showing results for just VALOR), with complete findings in Appendix \ref{best_practices_appendix}. In Fig. \ref{bp-b}, we compare performance with and without embeddings, using a uniform context distribution, for several choices of $K$ (the number of contexts). We find that using embeddings consistently improves the speed and stability of training. Fig. \ref{bp-b} also illustrates that training with a uniform distribution becomes more challenging as $K$ increases. In Figs. \ref{bp-c} and \ref{bp-d}, we show that agents with the curriculum trick and embeddings achieve mastery on $K_{max}=64$ contexts substantially faster than the agents trained with uniform context distributions in Fig. \ref{bp-b}. As shown in Appendix \ref{best_practices_appendix}, these results are consistent across algorithms.

\textbf{Comparison Study of Qualitative Results}: In our comparison, we tried to assess whether variational option discovery algorithms learn an interesting set of behaviors. This is subjective and hard to measure, so we restricted ourselves to testing for behaviors which are easy to quantify or observe; we note that there is substantial room in this space for developing performance metrics, and consider this an important avenue for future research. 

We trained agents by VALOR, VIC, and DIAYN, with embeddings and $K=64$ contexts, with and without the curriculum trick. We evaluated the learned behaviors by measuring the following quantities: final $x$-coordinate for Cheetah, final distance from origin for Swimmer, final distance from origin for Ant, and number of $z$-axis rotations for Ant\footnote{Approximately the number of complete circles walked by the agent around the ground-fixed $z$-axis (but not necessarily around the origin).}. We present partial findings in Fig. \ref{qualitative} and complete results in Appendix \ref{comparison_full}. Our results confirm findings from prior work, including \citet{Eysenbach2018} and \citet{Florensa2017}: variational option discovery methods, in some MuJoCo environments, are able to find locomotion gaits that travel in a variety of speeds and directions. 
Results in Cheetah and Ant are particularly good by this measure; in Swimmer, fairly few behaviors actually travel any meaningful distance from the origin ($>3$ units), but it happens non-negligibly often.  All three algorithms produce similar results in the locomotion domains, although we do find slight differences: particularly, DIAYN is more prone than VALOR and VIC to learn behaviors like `attain target state,' where the target state is fixed and unmoving. Our DIAYN behaviors are overall less mobile than the results reported by \citet{Eysenbach2018}; we believe that this is due to qualitative differences in how entropy is maximized by the underlying RL algorithms (soft actor-critic vs. entropy-regularized policy gradients).

We find that the curriculum approach does not appear to change the diversity of behaviors discovered in any large or consistent way. It appears to slightly increase the ranges for Cheetah $x$-coorindate, while slightly decreasing the ranges for Ant final distance. Scrutinizing the X-Y traces for all learned modes, it seems (subjectively) that the curriculum approach causes agents to move more erratically (see Appendices D.11---D.14). We do observe a particularly interesting effect for robustness: the curriculum approach makes the distribution of scores more consistent between random seeds (for performances of all seeds separately, see Appendices D.3---D.10).

We also attempted to perform a baseline comparison of all three variational option discovery methods against an approach where we used random reward functions in place of a learned decoder; however, we encountered substantial difficulties in optimizing with random rewards. The details of these experiments are given in Appendix \ref{random_reward_section}. 

\begin{figure}
\centering
\begin{subfigure}{0.49\textwidth}
  \includegraphics[width=\textwidth]{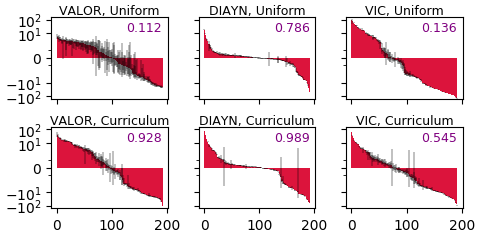}
  \caption{Final $x$-coordinate in Cheetah.}
\end{subfigure}
\begin{subfigure}{0.49\textwidth}
  \includegraphics[width=\textwidth]{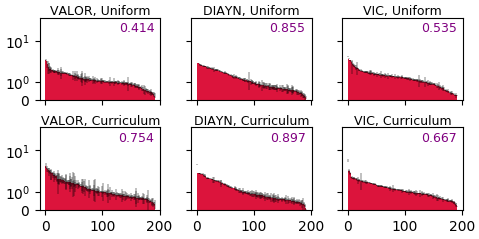}
  \caption{Final distance from origin in Swimmer.}
\end{subfigure}

\begin{subfigure}{0.49\textwidth}
  \includegraphics[width=\textwidth]{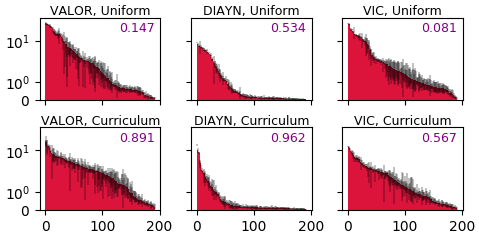}
  \caption{Final distance from origin in Ant.}
\end{subfigure}
\begin{subfigure}{0.49\textwidth}
  \includegraphics[width=\textwidth]{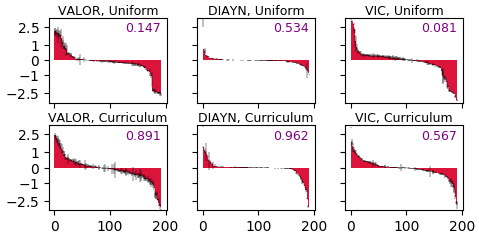}
  \caption{Number of $z$-axis rotations in Ant.}
\end{subfigure}
\caption{Bar charts illustrating scores for behaviors in Cheetah, Swimmer, and Ant, with $x$-axis showing behavior ID and $y$-axis showing the score in log scale. Each red bar (width $1$ on the $x$-axis) gives the average score for 5 trajectories conditioned on a single context; each chart is a composite from three random seeds, each of which was run with $K=64$ contexts, for a total of $192$ behaviors represented per chart. Behaviors were sorted in descending order by average score. Black bars show the standard deviation in score for a given behavior (context), and the upper-right corner of each chart shows the average decoder probability $\E[P_D(\tau|c)]$.}
\label{qualitative}
\end{figure}

\begin{figure}
\centering
\begin{subfigure}{0.21\textwidth}
  \includegraphics[width=\textwidth]{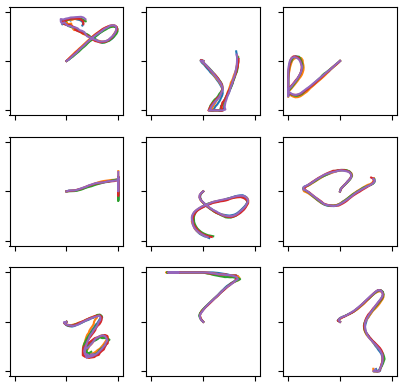}
  \caption{X-Y traces of example modes in Point.}
  \label{point-big-example}
\end{subfigure}\hspace{3mm}
\begin{subfigure}{0.21\textwidth}
  \includegraphics[width=\textwidth]{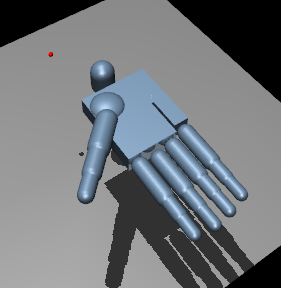}
  \caption{Robot hand environment.}
\end{subfigure}\hspace{3mm}
\begin{subfigure}{0.21\textwidth}
  \includegraphics[width=\textwidth]{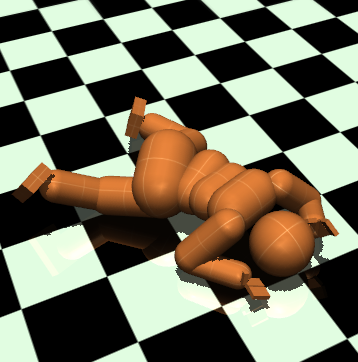}
  \caption{Toddler environment.}
\end{subfigure}\hspace{3mm}
\begin{subfigure}{0.21\textwidth}
  \includegraphics[width=\textwidth]{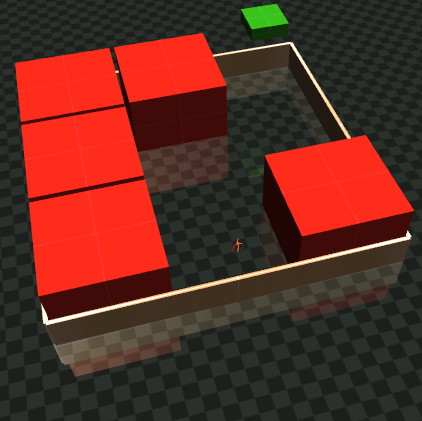}
  \caption{Ant-Maze environment.}
  \label{ant-maze}
\end{subfigure}

\begin{subfigure}{0.21\textwidth}
  \includegraphics[width=\textwidth]{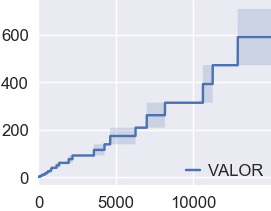}
  \caption{Point, current $K$.}
  \label{point-big-curve}
\end{subfigure}\hspace{3mm}
\begin{subfigure}{0.21\textwidth}
  \includegraphics[width=\textwidth]{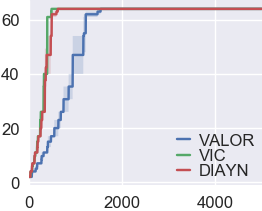}
  \caption{Hand, current $K$.}
  \label{hand-curves}
\end{subfigure}\hspace{3mm}
\begin{subfigure}{0.21\textwidth}
  \includegraphics[width=\textwidth]{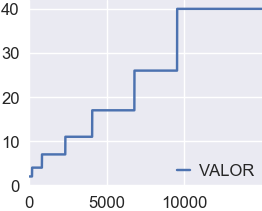}
  \caption{Toddler, current $K$.}
  \label{baby-curve}
\end{subfigure}\hspace{3mm}
\begin{subfigure}{0.21\textwidth}
  \includegraphics[width=\textwidth]{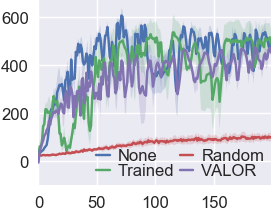}
  \caption{Ant-Maze return.}
  \label{ant-maze-curves}
\end{subfigure}

\caption{Various figures for spotlight experiments. Figs. \ref{point-big-example} and \ref{point-big-curve} show results from learning hundreds of behaviors in the Point env, with $K_{max} = 1024$. Fig. \ref{hand-curves} shows that optimizing Eq. \ref{vaerl} in the Hand environment is quite easy with the curriculum approach; all agents master the $K_{max}=64$ contexts in $<2000$ iterations. Fig. \ref{baby-curve} illustrates the challenge for variational option discovery in Toddler: after $15000$ iterations, only $K=40$ behaviors have been learned. Fig. \ref{ant-maze} shows the Ant-Maze environment, where red obstacles prevent the ant from reaching the green goal. Fig. \ref{ant-maze-curves} shows performance in Ant-Maze for different choices of a low-level policy in a hierarchy; in the Random and VALOR experiments, the low-level policy receives no gradient updates.}
\label{spotlight}
\end{figure}

\textbf{Hand and Toddler Environments}: Optimizing in the Hand environment (Fig. \ref{hand-curves}) was fairly easy and usually produced some naturalistic behaviors (eg pointing, bringing thumb and forefinger together, and one common rude gesture) as well as various unnatural behaviors (hand splayed out in what would be painful poses). Optimizing in the Toddler environment (Fig. \ref{baby-curve}) was highly challenging; the agent frequently struggled to learn more than a handful of behaviors. The behaviors which the agent did learn were extremely unnatural. We believe that this is because of a fundamental limitation of purely information-theoretic RL objectives: humans have strong priors on what constitutes natural behavior, but for sufficiently complex systems, those behaviors form a set of measure zero in the space of all possible behaviors; when a purely information-theoretic objective function is used, it will give no preference to the behaviors humans consider natural. 

\textbf{Learning Hundreds of Behaviors}: Via the curriculum approach, we are able to train agents in the Point environment to learn hundreds of behaviors which are distinct according to the decoder (Fig. \ref{point-big-curve}). We caution that this does not necessarily expand the space of behaviors which are learnable---it may merely allow for increasingly fine-grained binning of already-learned behaviors into contexts. From various experiments prior to our final results, we developed an intuition that it was important to carefully consider the capacity of the decoder here: the greater the decoder's capacity, the more easily it would overfit to undetectably-small differences in trajectories.  

\begin{figure}
\centering

\begin{subfigure}{0.45\textwidth}
  \includegraphics[width=\textwidth]{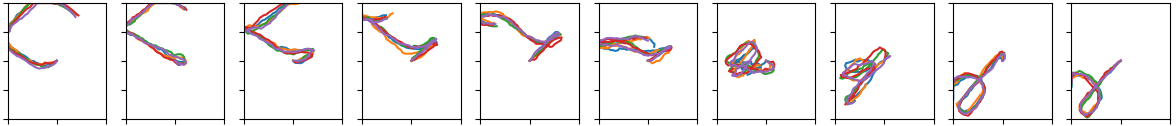}
  \caption{Interpolating behavior in the point environment.}
\end{subfigure}\hspace{5mm}%
\begin{subfigure}{0.45\textwidth}
  \includegraphics[width=\textwidth]{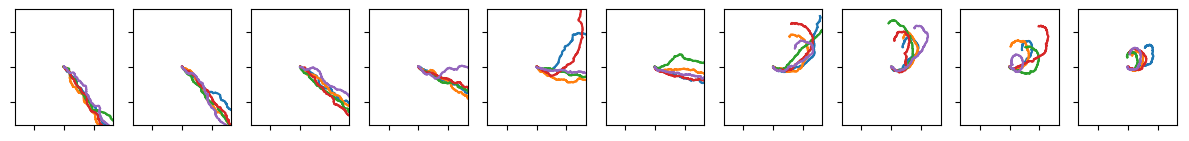}
  \caption{Interpolating behavior in the ant environment.}
\end{subfigure}
\caption{Plots on the far left and far right show X-Y traces for behaviors learned by VALOR; in-between plots show the X-Y traces conditioned on interpolated contexts.}
\label{interp}
\end{figure}

\textbf{Mode Interpolation}: We experimented with interpolating between context embeddings for point and ant policies to see if we could obtain interpolated behaviors. As shown in Fig. \ref{interp}, we found that some reasonably smooth interpolations were possible. This suggests that even though only a discrete number of behaviors are trained, the training procedure learns general-purpose universal policies.

\textbf{Downstream Tasks}: We investigated whether behaviors learned by variational option discovery could be used for a downstream task by taking a policy trained with VALOR on the Ant robot (Uniform distribution, seed 10; see Appendix D.7), and using it as the lower level of a two-level hierarchical policy in Ant-Maze. We held the VALOR policy fixed throughout downstream training, and only trained the upper level policy, using A2C as the RL algorithm (with reinforcement occuring only at the lower level---the upper level actions were trained by signals backpropagated through the lower level). Results are shown in Fig. \ref{ant-maze-curves}. We compared the performance of the VALOR-based agent to three baselines: a hierarchical agent with the same architecture trained from scratch on Ant-Maze (`Trained' in Fig. \ref{ant-maze-curves}), a hierarchical agent with a fixed random network as the lower level (`Random' in Fig. \ref{ant-maze-curves}), and a non-hierarchical agent with the same architecture as the upper level in the hierarchical agents (an MLP with one hidden layer, `None' in Fig. \ref{ant-maze-curves}). We found that the VALOR agent worked as well as the hierarchy trained from scratch and the non-hierarchical policy, with qualitatively similar learning curves for all three; the fixed random network performed quite poorly by comparison. This indicates that the space of options learned by (the particular run of) VALOR was at least as expressive as primitive actions, for the purposes of the task, and that VALOR options were more expressive than random networks here.

\section{Conclusions}

We performed a thorough empirical examination of variational option discovery techniques, and found they produce interesting behaviors in a variety of environments (such as Cheetah, Ant, and Hand), but can struggle in very high-dimensional control, as shown in the Toddler environment. From our mode interpolation and hierarchy experiments, we found evidence that the learned policies are universal in meaningful ways; however, we did not find clear evidence that hierarchies built on variational option discovery would outperform task-specific policies learned from scratch. 

We found that with purely information-theoretic objectives, agents in complex environments will discover behaviors that encode the context in trivial ways---eg through tiling a narrow volume of the state space with contexts. Thus a key challenge for future variational option discovery algorithms is to make the decoder distinguish between trajectories in a way which corresponds with human intuition about meaningful differences.

\subsubsection*{Acknowledgments}

Joshua Achiam is supported by TRUST (Team for Research in Ubiquitous Secure Technology) which receives support from NSF (award number CCF-0424422).
\bibliographystyle{plainnat}
\bibliography{valor_bib.bib}

\appendix

\section{VAE-Equivalence Proof}

The KL-divergence of $P(\tau|\pi,c)$ from $P(\tau| \pi_0)$ is
\begin{align*}
D_{KL}\left( P(\tau|\pi,c) || P(\tau | \pi_0) \right) &= \underE{\tau \sim \pi, c}{ \log \frac{P(\tau|\pi,c)}{P(\tau | \pi_0)}} \\
&= \underE{\tau \sim \pi, c}{ \log \frac{\mu(s_0) \prod_{t=0}^{T-1} P(s_{t+1}|s_t, a_t) \pi(a_t | s_t, c)}{\mu(s_0) \prod_{t=0}^{T-1} P(s_{t+1}|s_t, a_t) \pi_0 (a_t | s_t)}} \\
&= \underE{\tau \sim \pi, c}{ \sum_{t=0}^{T-1} \log \pi(a_t | s_t, c) - \log \pi_0 (a_t |s_t) } \\
&= -\calH(\pi,c) - \underE{\tau \sim \pi,c}{ \sum_{t=0}^{T-1} \log \pi_0 (a_t | s_t)}.
\end{align*}
The first term is our entropy regularization term. The second term, for a uniform random policy $\pi_0$, is a constant independent of $\pi$ (as long as $T$ is the same for all episodes) and can thus be removed from the objective function without changing the optimization problem.

\section{Implementation Details}

\subsection{Policy Optimization Algorithm} \label{implementation}

In this section, we will describe how we performed policy optimization for our experiments. We used vanilla policy gradient to optimize the reinforcement objective for all three variational option discovery algorithms,
\begin{equation*}
\nabla_{\theta} J(\pi_\theta) = \underE{c \sim G \\ \tau \sim \pi, c}{\sum_{t=0}^T \nabla_{\theta} \log \pi_{\theta} (a_t|s_t, c) \hat{A}_t},
\end{equation*}
although details varied slightly between algorithms and environments. The variation between environments was due to the presence or absence of extrinsic rewards. In all environments except for Ant, there were no extrinsic rewards; however, in Ant, a small penalty was applied for falling over (as opposed to terminating the episode when the agent falls over, as in \citet{Eysenbach2018}). 

\begin{itemize}
\item For VALOR and VIC, the advantage function was:
\begin{equation*}
\hat{A}_t = \text{normalize}\left(\log P_D (c|\tau)\right) + \text{normalize}\left(\sum_{t'=t}^T \left(\gamma^{t'-t} r_{t'} - V_{\psi} (s_t, c)\right)\right),
\end{equation*}
where the $\text{normalize}$ function subtracts out the batch mean and divides by the batch standard deviation, and $V_{\psi}$ was a learned value function baseline. $V_{\psi} (s_t, c)$ was learned by taking one gradient descent step on
\begin{equation*}
\min_{\psi} \sum_{ (s_t, c) \in \calD} \left(\gamma^{t'-t} r_{t'} - V_{\psi} (s_t, c)\right)^2
\end{equation*}
per iteration.
\item For DIAYN, the advantage function was:
\begin{equation*}
\hat{A}_t = \text{normalize}\left( \sum_{t'=t}^T \left(\gamma^{t'-t} \left(\log P_D (c|s_{t'}) + r_{t'} \right) - V_{\psi} (s_t, c)\right)\right)
\end{equation*}
where $V_{\psi} (s_t, c)$ was learned by descending on 
\begin{equation*}
\min_{\psi} \sum_{ (s_t, c) \in \calD} \left(\gamma^{t'-t} \left(\log P_D (c|s_{t'}) + r_{t'}\right) - V_{\psi} (s_t, c)\right)^2.
\end{equation*}
\end{itemize}

When computing the gradient of the entropy term, we made an approximation that ignored the role of $\pi$ in the distribution over trajectories:
\begin{align*}
\nabla_{\theta} \calH(\pi,c) &= \nabla_{\theta} \sum_{t=0}^{T-1} \underE{s_t \sim \pi, c}{H(\pi(\cdot|s_t, c))} \\
&\approx \sum_{t=0}^{T-1} \underE{s_t \sim \pi, c}{\nabla_{\theta}H(\pi(\cdot|s_t, c))},
\end{align*}
resulting in the same entropy regularization as in \citet{Mnih2016}. Following practices for vanilla policy gradient established in \cite{Duan2016}, we use the Adam optimizer \cite{Kingma2015}.

\subsection{Hyperparameters}

For all variational option discovery algorithms, we used:
\begin{itemize}
\item $1000$ paths per epoch for the policy gradient batch
\item $\gamma=0.97$ as the discount factor
\item $\beta=1e^{-3}$ as the entropy regularization coefficient, where applicable (omitted for VIC)
\item $1e^{-3}$ as the Adam learning rate
\item LSTM($64$) followed by MLP($32$) with $\tanh$ activations as the policy architecture
\item $32$ as the context embedding dimension (when using context embeddings)
\end{itemize}

For VALOR, the decoder was a bidirectional LSTM where the cell for each direction was of size $64$. For VIC and DIAYN, the decoder was an MLP of size $(180,180)$. 

\newpage



\section{Additional Analysis for Best Practices} \label{best_practices_appendix}
\FloatBarrier
\begin{figure}[H]
\centering

\rule{0.75\linewidth}{0.5pt}

VALOR: \vspace{2mm}

\begin{subfigure}{.24\textwidth}
  \includegraphics[height=2.4cm]{images/best_practices3/cheetah_valor_num_modes_and_embed_ablation.png}
  \caption{Uniform, for various $K$, $\log P_D$}
  \label{bp-valor-b}
\end{subfigure}%
\begin{subfigure}{.24\textwidth}
  \includegraphics[height=2.4cm]{images/best_practices3/cheetah_valor_c64_curriculum_vs_uniform.png}  
  \caption{Uniform vs Curriculum, $\log P_D$}
  \label{bp-valor-c}
\end{subfigure}%
\begin{subfigure}{.24\textwidth}
  \includegraphics[height=2.4cm]{images/best_practices3/cheetah_valor_c64_curriculum_current_n.png}  
  \caption{Curriculum, current $K$ \newline}
  \label{bp-valor-d}
\end{subfigure}

\vspace{2mm}

\rule{0.75\linewidth}{0.5pt}

VIC: \vspace{2mm}

\begin{subfigure}{.24\textwidth}
  \includegraphics[height=2.4cm]{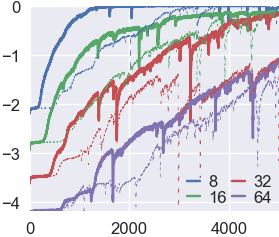}
  \caption{Uniform, for various $K$, $\log P_D$}
  \label{bp-vic-b}
\end{subfigure}%
\begin{subfigure}{.24\textwidth}
  \includegraphics[height=2.4cm]{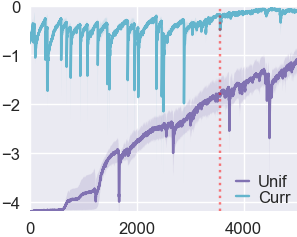}  
  \caption{Uniform vs Curriculum, $\log P_D$}
  \label{bp-vic-c}
\end{subfigure}%
\begin{subfigure}{.24\textwidth}
  \includegraphics[height=2.4cm]{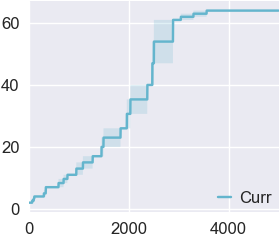}  
  \caption{Curriculum, current $K$ \newline}
  \label{bp-vic-d}
\end{subfigure}

\rule{0.75\linewidth}{0.5pt}

DIAYN: \vspace{2mm}

\begin{subfigure}{.24\textwidth}
  \includegraphics[height=2.4cm]{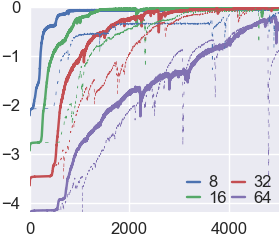}
  \caption{Uniform, for various $K$, $\log P_D$}
  \label{bp-diayn-b}
\end{subfigure}%
\begin{subfigure}{.24\textwidth}
  \includegraphics[height=2.4cm]{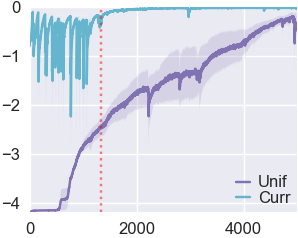}  
  \caption{Uniform vs Curriculum, $\log P_D$}
  \label{bp-diayn-c}
\end{subfigure}%
\begin{subfigure}{.24\textwidth}
  \includegraphics[height=2.4cm]{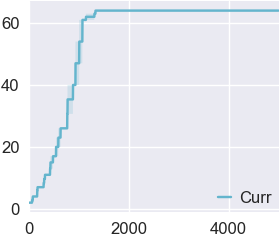}  
  \caption{Curriculum, current $K$ \newline}
  \label{bp-diayn-d}
\end{subfigure}

\caption[]{Analysis for understanding best training practices for various algorithms with HalfCheetah as the environment. The $x$-axis is number of training iterations, and in (a) and (b), the $y$-axis is $\E[\log P_D(c|\tau)]$; in (c), the $y$-axis gives the current value of $K$ in the curriculum. (a) shows a direct comparison between learning curves with (dark) and without (dotted) context embeddings, for $K\in\{8,16,32,64\}$. (b) shows learning performance for the curriculum approach with $K_{max}=64$, compared against the uniform distribution approach with $K=64$: the spikes and dips for the curriculum curve are characteristic of points when $K$ changes according to Eq. \ref{curriculum}. The dashed red line shows when $K=K_{max}$ for the curriculum approach; prior to it, the curves for Uniform and Curriculum are not directly comparable, but after it, they are. (c) shows $K$ for the curriculum approach throughout the runs from (b). All curves are averaged over three random seeds.}
\label{best_practices_all}
\end{figure}

\newpage

\section{Complete Experimental Results for Comparison Study} \label{comparison_full}

\subsection{Guide to Reading This Section}

In this section we present the results from our core comparison of \{VALOR, VIC, DIAYN\} $\times$ \{Uniform, Curriculum\}. Because these algorithms perform unsupervised behavior discovery, analyzing our results is highly-challenging: there is no single, quantitative measure by which to compare the algorithms. We choose to examine our results in a variety of ways:
\begin{itemize}
\item Learning curves for the optimization objective.
\item Bar charts and histograms to show scores for the learned behaviors. Particularly, we evaluate final $x$-coordinate in the Cheetah environment, final distance traveled in the Swimmer environment, final distance traveled in the Ant environment, and number of $z$-axis rotations in the Ant environment. Scores are evaluated on trajectories of length $T=1000$ steps, even though agents are trained on trajectories with $T=250$; we find that using longer horizons at test time clarifies the differences between behaviors.
\item X-Y traces for agent trajectories in the Point and Ant environments. (X-Y traces for the center-of-mass in Swimmer are not very insightful: Swimmer behavior is highly oscillatory and so it is difficult to discern what is happening.) 
\end{itemize}

Regarding the bar charts and histograms in subsections D.3---D.10: 
\begin{itemize}
\item The bar charts are arranged in nearly the same way as the charts in \ref{qualitative}: the $x$-axis is behavior ID, and the $y$-axis shows score in log scale for that behavior. The black bars show standard deviations for behavior scores.
\item The histograms show score on the $x$-axis, and number of behaviors that fall into a given bin on the $y$-axis in log scale. 
\item The charts for `all' show the composite bars for all behaviors from seeds $0$, $10$, and $20$. The `s0', `s10', and `s20' charts show behaviors from particular random seeds. Each single seed corresponds to a single policy with $K=64$ behaviors.
\end{itemize}

Regarding the X-Y traces in subsections D.11---D.14: 
\begin{itemize}
\item In the Point traces, the ranges for $x$ and $y$ are $x\in [-1.3, 1.3]$ and $y \in [-1.3, 1.3]$.
\item In the Ant traces, the ranges for $x$ and $y$ are $x\in[-15,15]$ and $y \in [-15, 15]$.
\item For the Point environment, traces are taken from trajectories with the same time horizon as training ($T=65$); for the Ant environment, we use the $T=1000$ trajectories.
\end{itemize}

\newpage
\subsection{Learning Curves} \label{subsec_learning_curves}

\begin{figure}[h]
\centering

\rule{0.75\linewidth}{0.5pt}

Point Env: \vspace{2mm}

\begin{subfigure}{.24\textwidth}
  \centering
  \includegraphics[width=\linewidth]{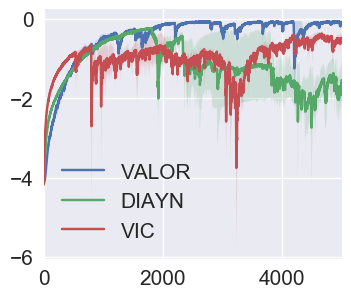}
  \caption{$\log P_D$, Uniform}
\end{subfigure}\hspace{5mm}%
\begin{subfigure}{.24\textwidth}
  \centering
  \includegraphics[width=\linewidth]{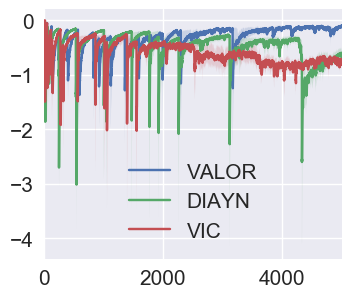}
  \caption{$\log P_D$, Curriculum}
\end{subfigure}\hspace{5mm}%
\begin{subfigure}{.24\textwidth}
  \centering
  \includegraphics[width=\linewidth]{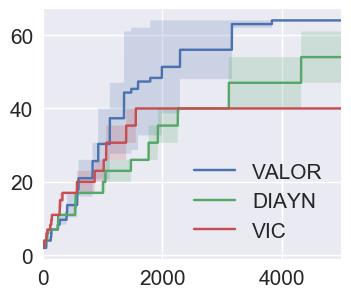}
  \caption{$K_{cur}$, Curriculum}
\end{subfigure}

\rule{0.75\linewidth}{0.5pt}

Cheetah: \vspace{2mm}

\begin{subfigure}{.24\textwidth}
  \centering
  \includegraphics[width=\linewidth]{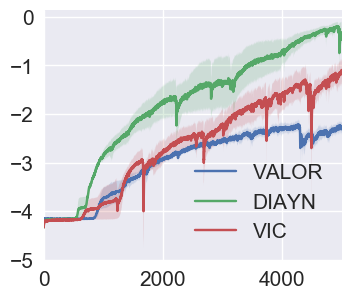}
  \caption{$\log P_D$, Uniform}
\end{subfigure}\hspace{5mm}%
\begin{subfigure}{.24\textwidth}
  \centering
  \includegraphics[width=\linewidth]{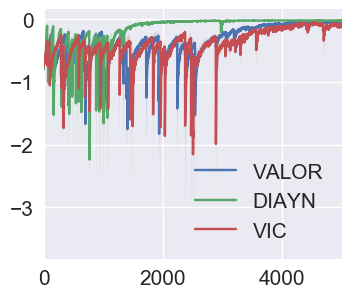}
  \caption{$\log P_D$, Curriculum}
\end{subfigure}\hspace{5mm}%
\begin{subfigure}{.24\textwidth}
  \centering
  \includegraphics[width=\linewidth]{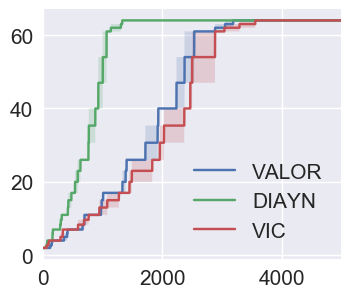}
  \caption{$K_{cur}$, Curriculum}
\end{subfigure}%

\rule{0.75\linewidth}{0.5pt}

Swimmer: \vspace{2mm}

\begin{subfigure}{.24\textwidth}
  \centering
  \includegraphics[width=\linewidth]{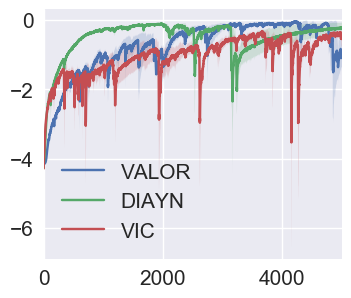}
  \caption{$\log P_D$, Uniform}
\end{subfigure}\hspace{5mm}%
\begin{subfigure}{.24\textwidth}
  \centering
  \includegraphics[width=\linewidth]{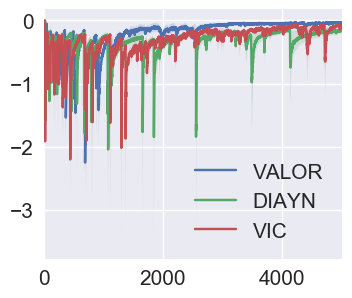}
  \caption{$\log P_D$, Curriculum}
\end{subfigure}\hspace{5mm}%
\begin{subfigure}{.24\textwidth}
  \centering
  \includegraphics[width=\linewidth]{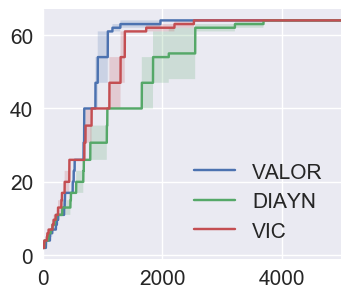}
  \caption{$K_{cur}$, Curriculum}
\end{subfigure}%

\rule{0.75\linewidth}{0.5pt}

Ant: \vspace{2mm}

\begin{subfigure}{.24\textwidth}
  \centering
  \includegraphics[width=\linewidth]{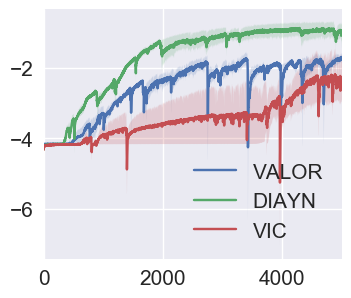}
  \caption{$\log P_D$, Uniform}
\end{subfigure}\hspace{5mm}%
\begin{subfigure}{.24\textwidth}
  \centering
  \includegraphics[width=\linewidth]{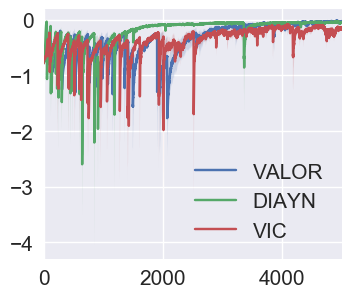}
  \caption{$\log P_D$, Curriculum}
\end{subfigure}\hspace{5mm}%
\begin{subfigure}{.24\textwidth}
  \centering
  \includegraphics[width=\linewidth]{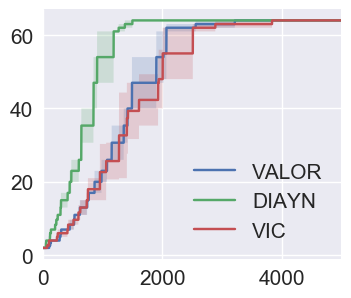}
  \caption{$K_{cur}$, Curriculum}
\end{subfigure}%

\caption[]{Learning curves for all algorithms and environments in our core comparison, for number of contexts $K=64$. The curriculum trick generally tends to speed up and stabilize performance, except for DIAYN and VIC in the point environment.}
\label{learning_curves}
\end{figure}

\FloatBarrier
\newpage
\subsection{Evaluating Learned Behaviors: Cheetah, Uniform Context Distribution}

\begin{figure}[h]
\centering
\rule{\linewidth}{0.5pt}
VALOR, Uniform Context Distribution: \vspace{2mm}
\includegraphics[width=\linewidth]{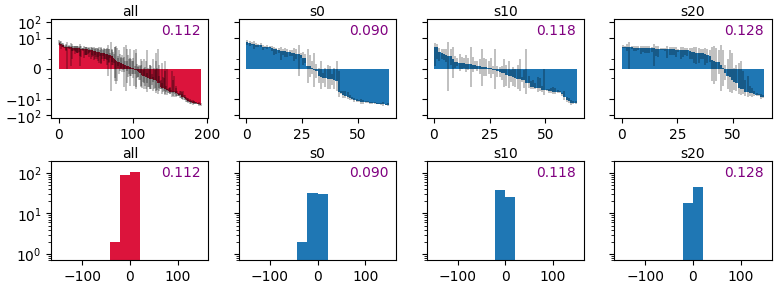}
\rule{\linewidth}{0.5pt}
VIC, Uniform Context Distribution: \vspace{2mm}
\includegraphics[width=\linewidth]{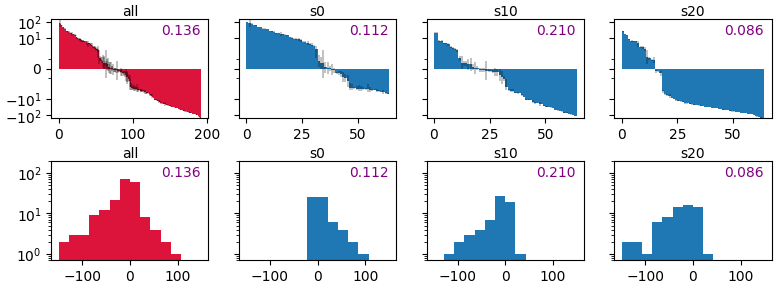}
\rule{\linewidth}{0.5pt}
DIAYN, Uniform Context Distribution: \vspace{2mm}
\includegraphics[width=\linewidth]{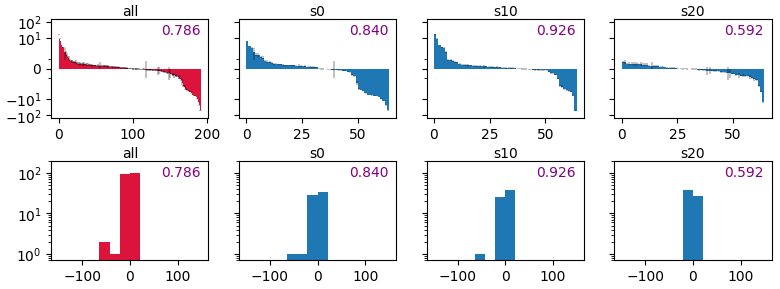}
\caption{Final $x$-coordinate in the Cheetah environment.}
\end{figure}

\FloatBarrier
\newpage
\subsection{Evaluating Learned Behaviors: Cheetah, Curriculum Context Distribution}

\begin{figure}[h]
\centering
\rule{\linewidth}{0.5pt}
VALOR, Curriculum Context Distribution: \vspace{2mm}
\includegraphics[width=\linewidth]{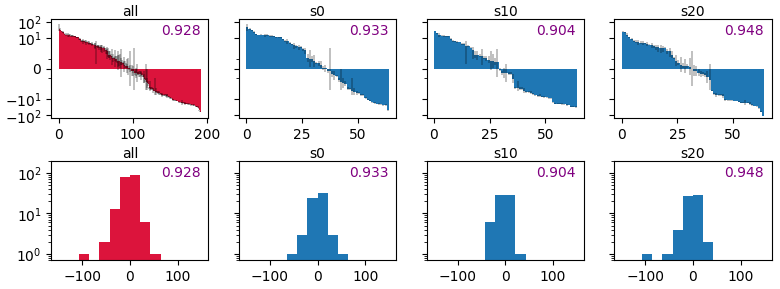}
\rule{\linewidth}{0.5pt}
VIC, Curriculum Context Distribution: \vspace{2mm}
\includegraphics[width=\linewidth]{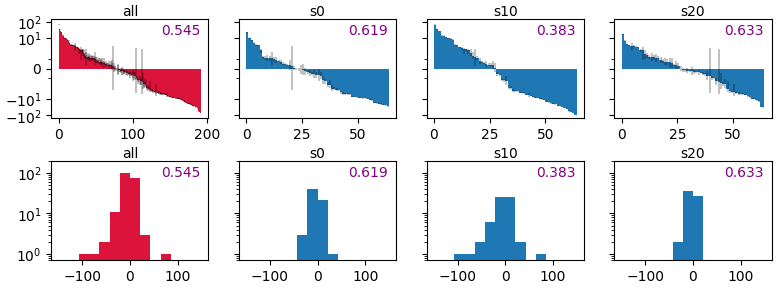}
\rule{\linewidth}{0.5pt}
DIAYN, Curriculum Context Distribution: \vspace{2mm}
\includegraphics[width=\linewidth]{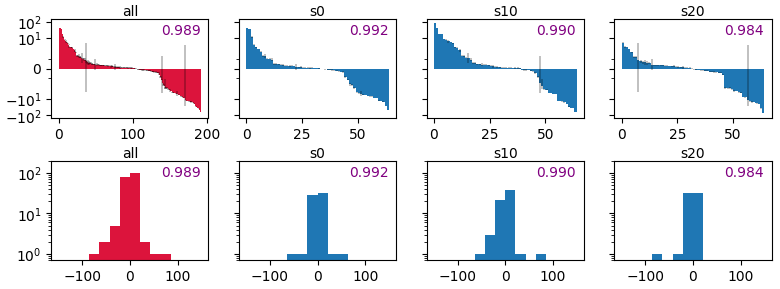}
\caption{Final $x$-coordinate in the Cheetah environment.}
\end{figure}

\FloatBarrier
\newpage
\subsection{Evaluating Learned Behaviors: Swimmer, Uniform Context Distribution}

\begin{figure}[h]
\centering
\rule{\linewidth}{0.5pt}
VALOR, Uniform Context Distribution: \vspace{2mm}
\includegraphics[width=\linewidth]{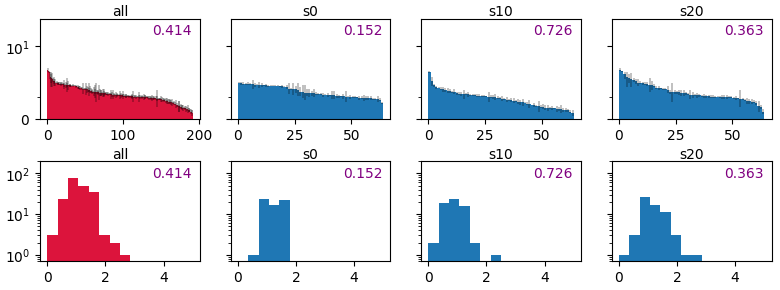}
\rule{\linewidth}{0.5pt}
VIC, Uniform Context Distribution: \vspace{2mm}
\includegraphics[width=\linewidth]{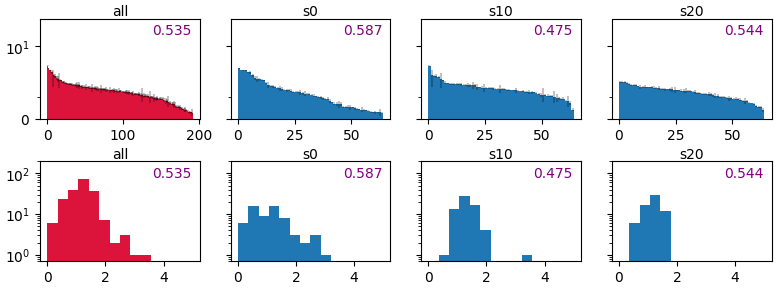}
\rule{\linewidth}{0.5pt}
DIAYN, Uniform Context Distribution: \vspace{2mm}
\includegraphics[width=\linewidth]{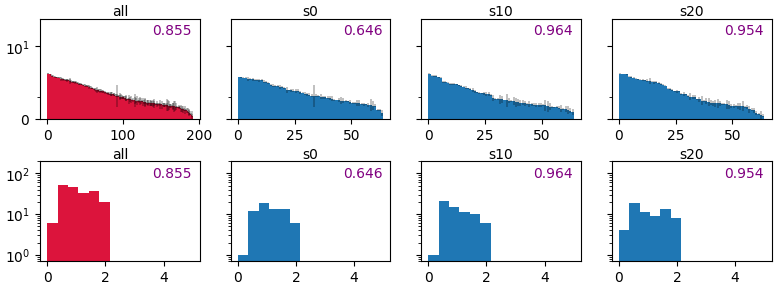}
\caption{Final distance from origin in the Swimmer environment.}
\end{figure}

\FloatBarrier
\newpage
\subsection{Evaluating Learned Behaviors: Swimmer, Curriculum Context Distribution}

\begin{figure}[h]
\centering
\rule{\linewidth}{0.5pt}
VALOR, Curriculum Context Distribution: \vspace{2mm}
\includegraphics[width=\linewidth]{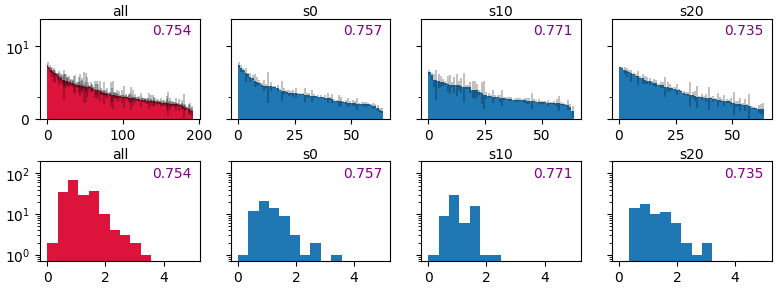}
\rule{\linewidth}{0.5pt}
VIC, Curriculum Context Distribution: \vspace{2mm}
\includegraphics[width=\linewidth]{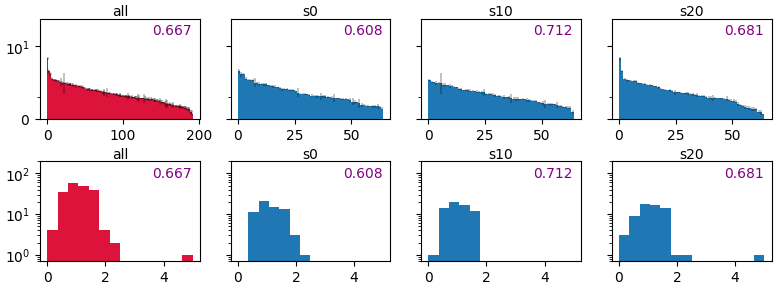}
\rule{\linewidth}{0.5pt}
DIAYN, Curriculum Context Distribution: \vspace{2mm}
\includegraphics[width=\linewidth]{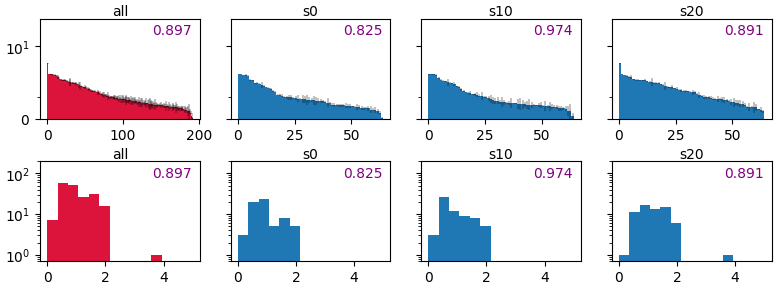}
\caption{Final distance from origin in the Swimmer environment.}
\end{figure}

\FloatBarrier
\newpage
\subsection{Evaluating Learned Behaviors: Ant (Distance), Uniform Context Distribution}

\begin{figure}[h]
\centering
\rule{\linewidth}{0.5pt}
VALOR, Uniform Context Distribution: \vspace{2mm}
\includegraphics[width=\linewidth]{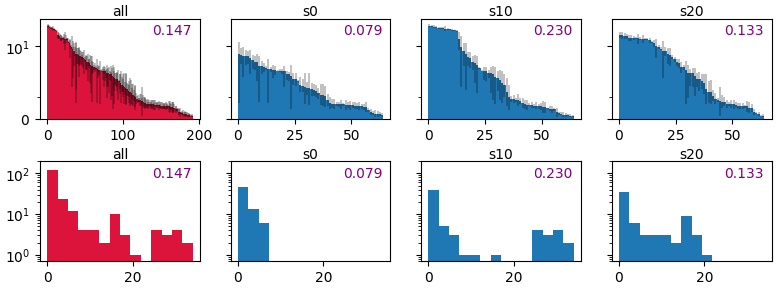}
\rule{\linewidth}{0.5pt}
VIC, Uniform Context Distribution: \vspace{2mm}
\includegraphics[width=\linewidth]{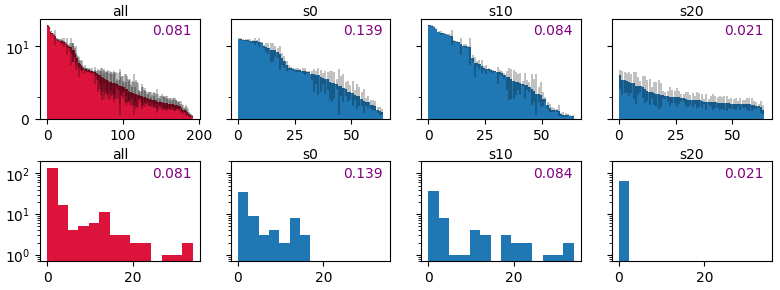}
\rule{\linewidth}{0.5pt}
DIAYN, Uniform Context Distribution: \vspace{2mm}
\includegraphics[width=\linewidth]{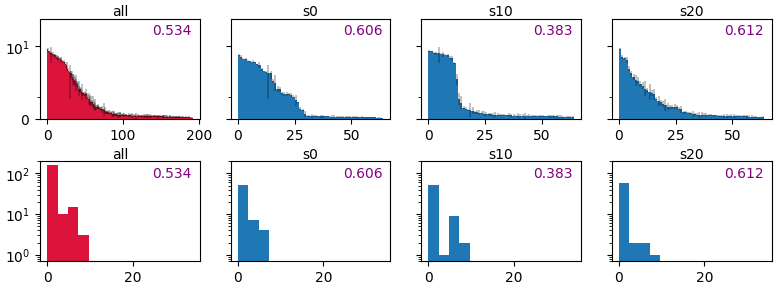}
\caption{Final distance from origin in the Ant environment.}
\end{figure}

\FloatBarrier
\newpage
\subsection{Evaluating Learned Behaviors: Ant (Distance), Curriculum Context Distribution}

\begin{figure}[h]
\centering
\rule{\linewidth}{0.5pt}
VALOR, Curriculum Context Distribution: \vspace{2mm}
\includegraphics[width=\linewidth]{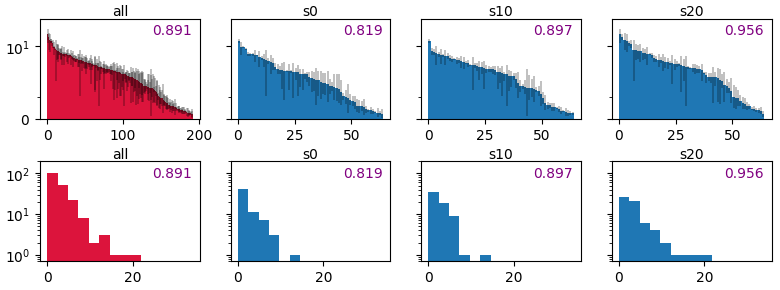}
\rule{\linewidth}{0.5pt}
VIC, Curriculum Context Distribution: \vspace{2mm}
\includegraphics[width=\linewidth]{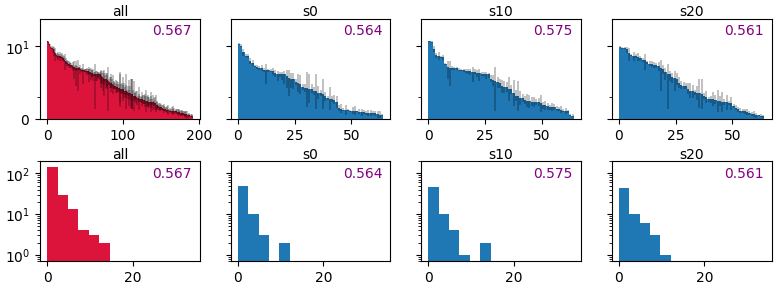}
\rule{\linewidth}{0.5pt}
DIAYN, Curriculum Context Distribution: \vspace{2mm}
\includegraphics[width=\linewidth]{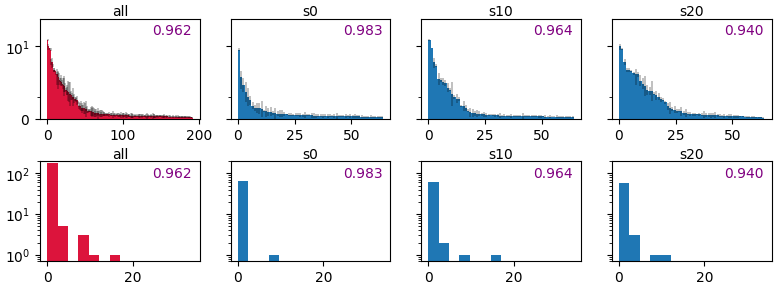}
\caption{Final distance from origin in the Ant environment.}
\end{figure}

\FloatBarrier
\newpage
\subsection{Evaluating Learned Behaviors: Ant (Rotations), Uniform Context Distribution}

\begin{figure}[h]
\centering
\rule{\linewidth}{0.5pt}
VALOR, Uniform Context Distribution: \vspace{2mm}
\includegraphics[width=\linewidth]{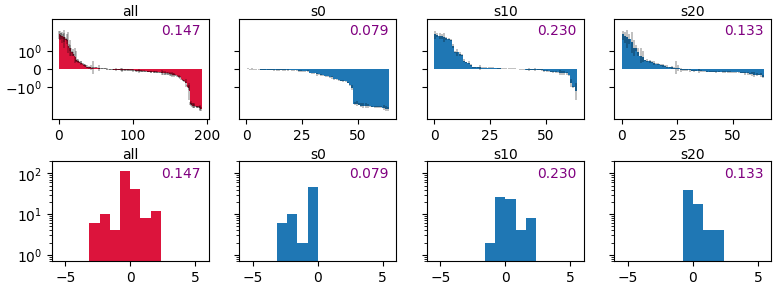}
\rule{\linewidth}{0.5pt}
VIC, Uniform Context Distribution: \vspace{2mm}
\includegraphics[width=\linewidth]{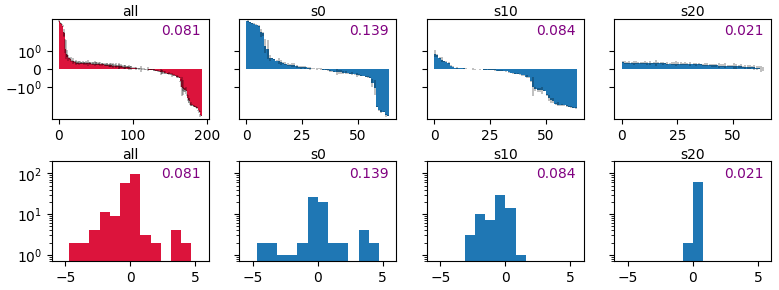}
\rule{\linewidth}{0.5pt}
DIAYN, Uniform Context Distribution: \vspace{2mm}
\includegraphics[width=\linewidth]{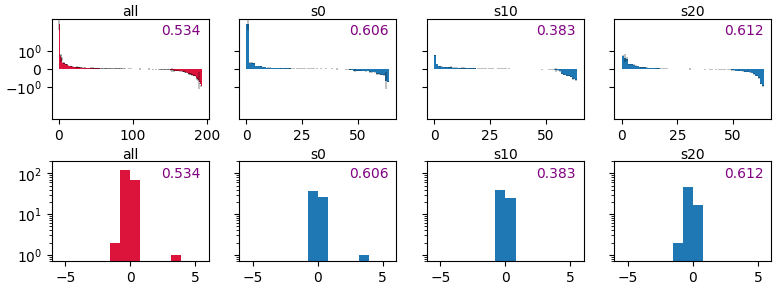}
\caption{Number of $z$-axis rotations in the Ant environment.}
\end{figure}

\FloatBarrier
\newpage
\subsection{Evaluating Learned Behaviors: Ant (Rotations), Curriculum Context Distribution}

\begin{figure}[h]
\centering
\rule{\linewidth}{0.5pt}
VALOR, Curriculum Context Distribution: \vspace{2mm}
\includegraphics[width=\linewidth]{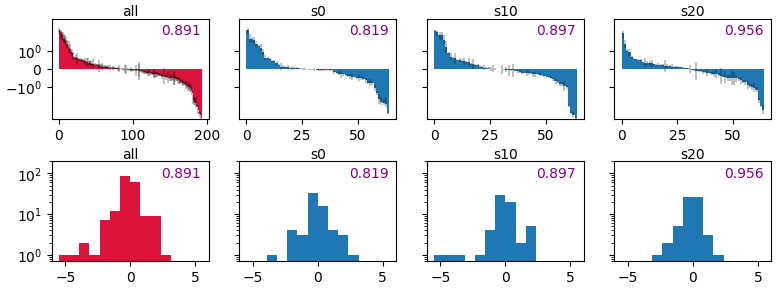}
\rule{\linewidth}{0.5pt}
VIC, Curriculum Context Distribution: \vspace{2mm}
\includegraphics[width=\linewidth]{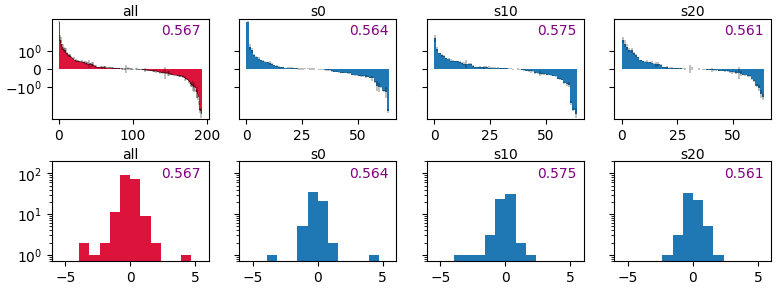}
\rule{\linewidth}{0.5pt}
DIAYN, Curriculum Context Distribution: \vspace{2mm}
\includegraphics[width=\linewidth]{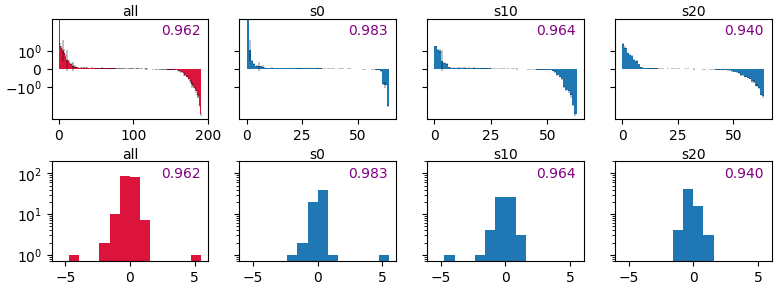}
\caption{Number of $z$-axis rotations in the Ant environment.}
\end{figure}

\FloatBarrier
\newpage
\subsection{Point Environment, Uniform Context Distribution, XY-Traces}
\FloatBarrier
\begin{figure}[h]
\centering

\rule{\linewidth}{0.5pt}
VALOR, Uniform Context Distribution:: \vspace{2mm}

\begin{subfigure}{0.32\textwidth}
  \includegraphics[width=\textwidth]{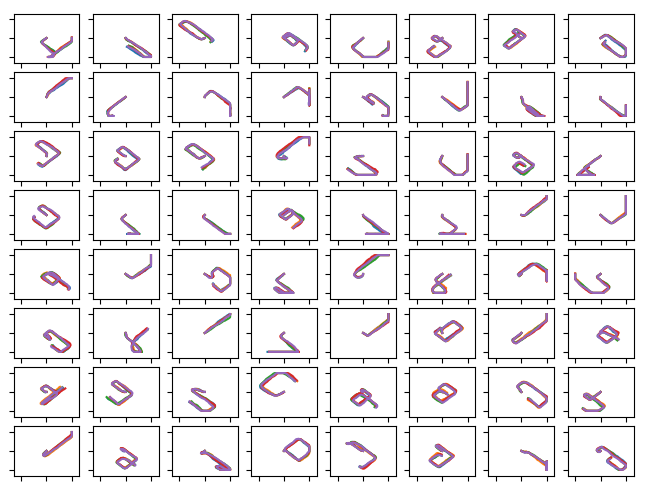}
  \caption{VALOR, Uniform, s0}
\end{subfigure}
\begin{subfigure}{0.32\textwidth}
  \includegraphics[width=\textwidth]{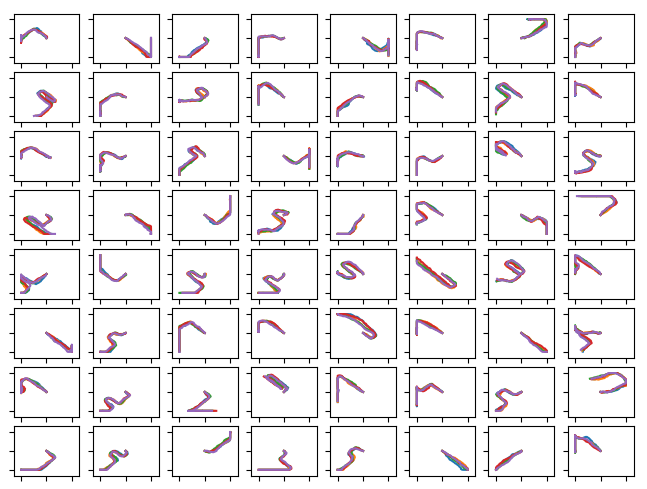}
  \caption{VALOR, Uniform, s10}
\end{subfigure}
\begin{subfigure}{0.32\textwidth}
  \includegraphics[width=\textwidth]{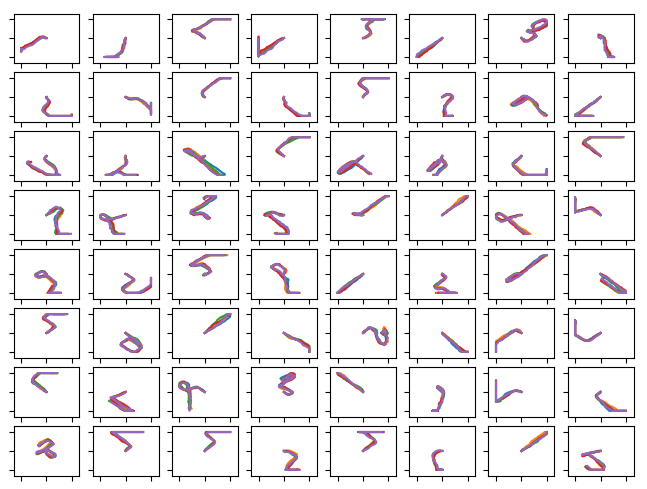}
  \caption{VALOR, Uniform, s20}
\end{subfigure}

\rule{\linewidth}{0.5pt}

VIC, Uniform Context Distribution:\vspace{2mm}

\begin{subfigure}{0.32\textwidth}
  \includegraphics[width=\textwidth]{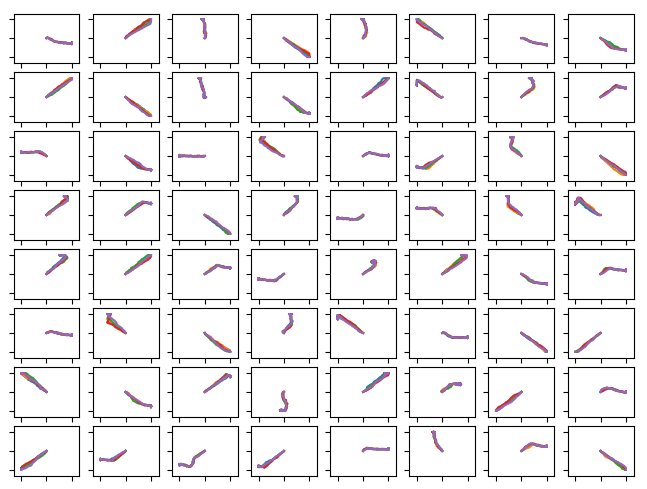}
  \caption{VIC, Uniform, s0}
\end{subfigure}
\begin{subfigure}{0.32\textwidth}
  \includegraphics[width=\textwidth]{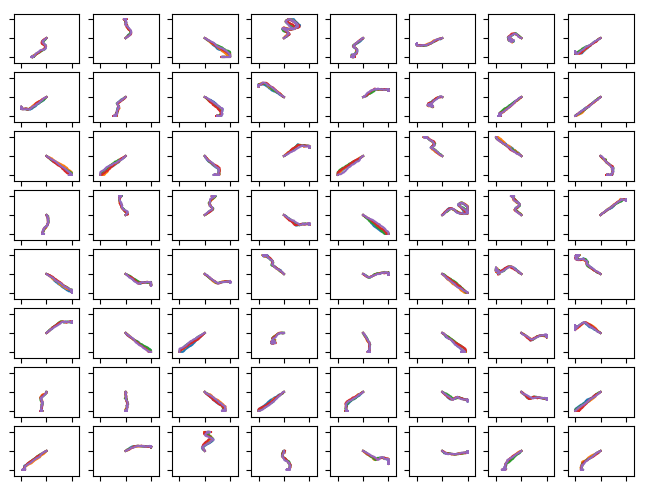}
  \caption{VIC, Uniform, s10}
\end{subfigure}
\begin{subfigure}{0.32\textwidth}
  \includegraphics[width=\textwidth]{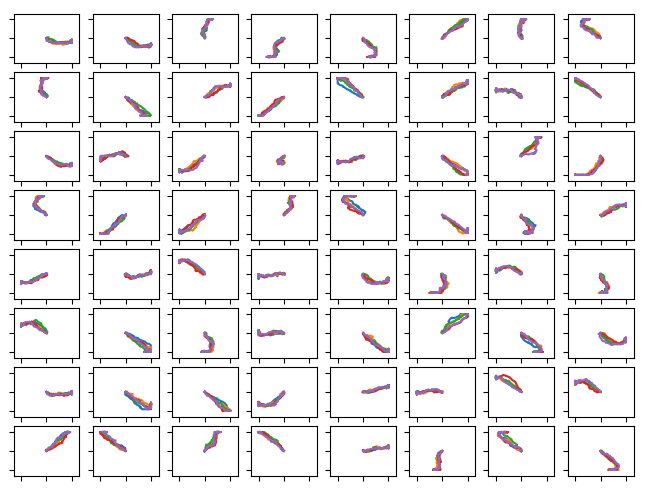}
  \caption{VIC, Uniform, s20}
\end{subfigure}

\rule{\linewidth}{0.5pt}

DIAYN, Uniform Context Distribution:\vspace{2mm}

\begin{subfigure}{0.32\textwidth}
  \includegraphics[width=\textwidth]{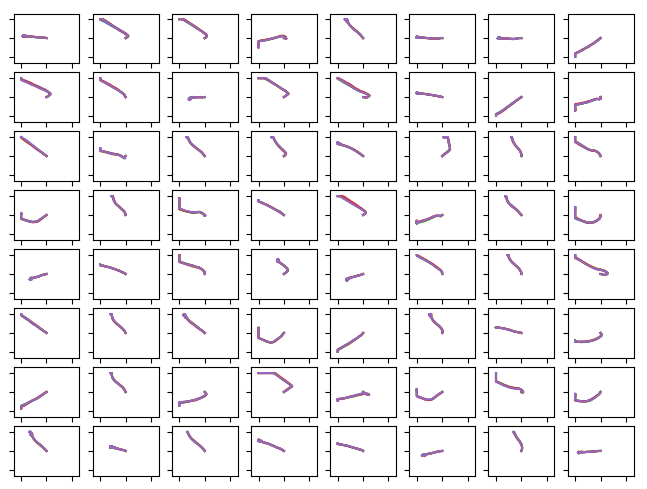}
  \caption{DIAYN, Uniform, s0}
\end{subfigure}
\begin{subfigure}{0.32\textwidth}
  \includegraphics[width=\textwidth]{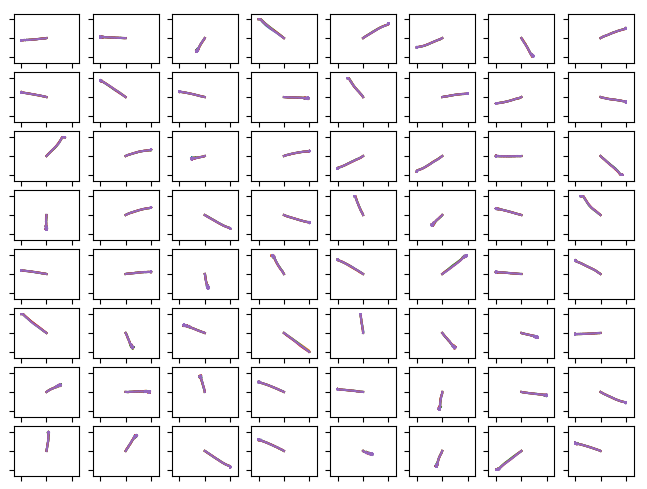}
  \caption{DIAYN, Uniform, s10}
\end{subfigure}
\begin{subfigure}{0.32\textwidth}
  \includegraphics[width=\textwidth]{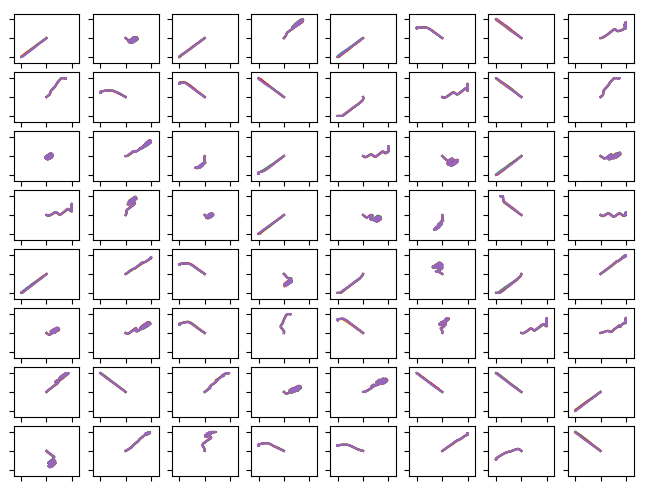}
  \caption{DIAYN, Uniform, s20}
\end{subfigure}

\caption{Learned behaviors in the Point environment with uniform context distributions. Each sub-plot shows X-Y traces for five trajectories conditioned on the same context (because the learned behaviors are highly repeatable, most traces almost entirely overlap). All traces for an algorithm come from a single policy which was trained with $K=64$ contexts.}
\label{xy_point_uniform}
\end{figure}

\newpage
\subsection{Point Environment, Curriculum Context Distribution, XY-Traces}
\FloatBarrier

\begin{figure}[h]
\centering

\rule{\linewidth}{0.5pt}
VALOR, Curriculum Context Distribution: \vspace{2mm}

\begin{subfigure}{0.32\textwidth}
  \includegraphics[width=\textwidth]{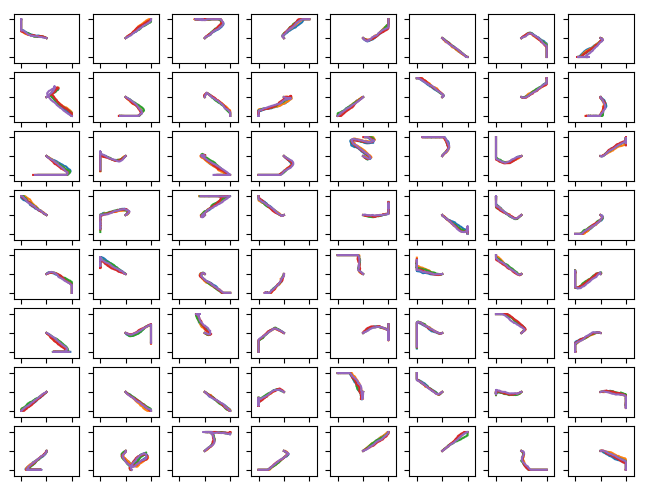}
  \caption{VALOR, Curriculum, s0}
\end{subfigure}
\begin{subfigure}{0.32\textwidth}
  \includegraphics[width=\textwidth]{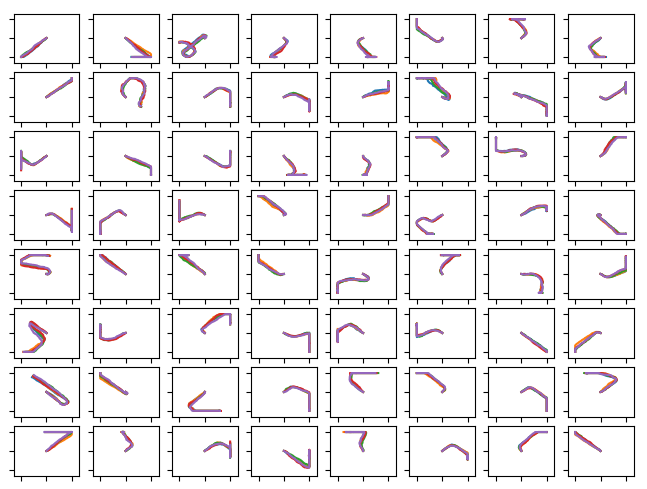}
  \caption{VALOR, Curriculum, s10}
\end{subfigure}
\begin{subfigure}{0.32\textwidth}
  \includegraphics[width=\textwidth]{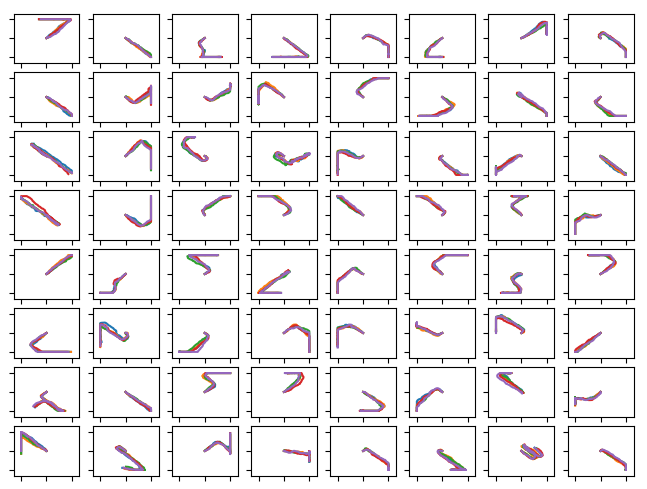}
  \caption{VALOR, Curriculum, s20}
\end{subfigure}

\rule{\linewidth}{0.5pt}

VIC, Curriculum Context Distribution: \vspace{2mm}

\begin{subfigure}{0.32\textwidth}
  \includegraphics[width=\textwidth]{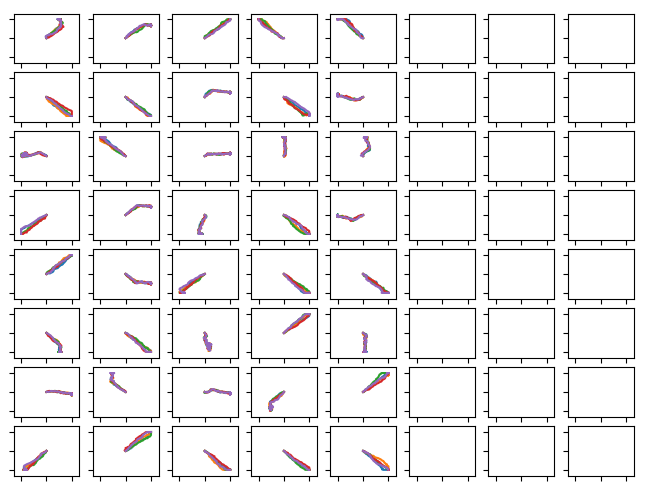}
  \caption{VIC, Curriculum, s0}
\end{subfigure}
\begin{subfigure}{0.32\textwidth}
  \includegraphics[width=\textwidth]{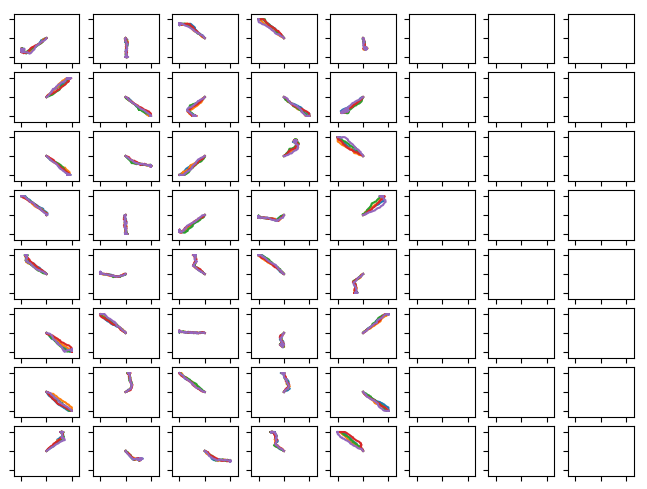}
  \caption{VIC, Curriculum, s10}
\end{subfigure}
\begin{subfigure}{0.32\textwidth}
  \includegraphics[width=\textwidth]{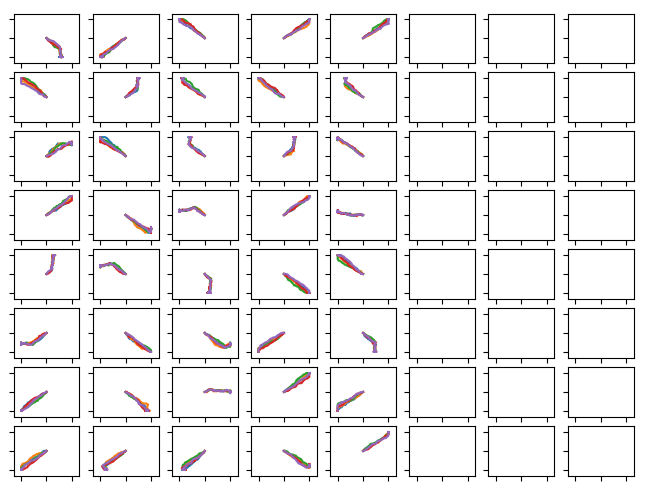}
  \caption{VIC, Curriculum, s20}
\end{subfigure}

\rule{\linewidth}{0.5pt}

DIAYN, Curriculum Context Distribution:\vspace{2mm}

\begin{subfigure}{0.32\textwidth}
  \includegraphics[width=\textwidth]{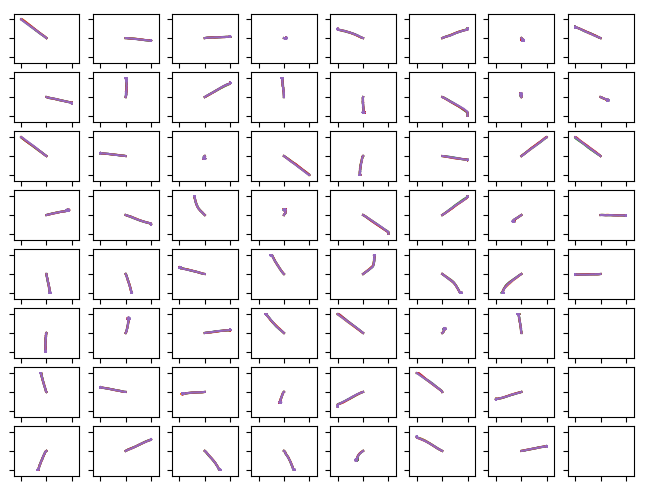}
  \caption{DIAYN, Curriculum, s0}
\end{subfigure}
\begin{subfigure}{0.32\textwidth}
  \includegraphics[width=\textwidth]{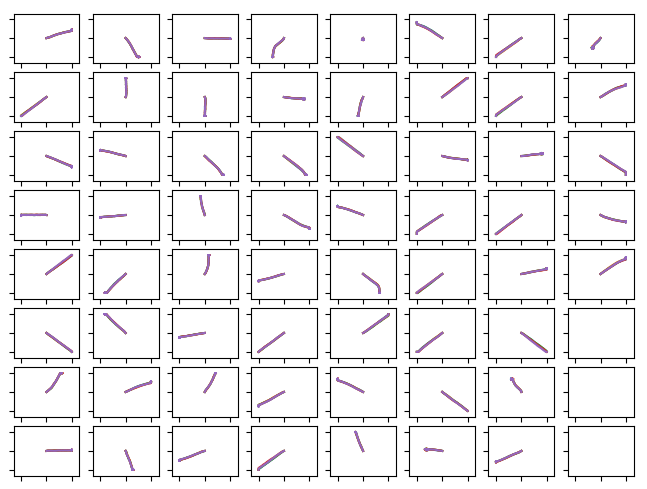}
  \caption{DIAYN, Curriculum, s10}
\end{subfigure}
\begin{subfigure}{0.32\textwidth}
  \includegraphics[width=\textwidth]{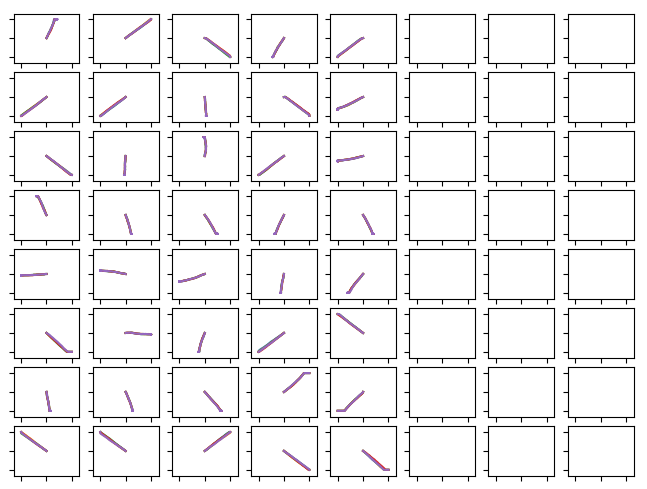}
  \caption{DIAYN, Curriculum, s20}
\end{subfigure}

\caption{Learned behaviors in the Point environment with the curriculum trick. Each sub-plot shows X-Y traces for five trajectories conditioned on the same context (because the learned behaviors are highly repeatable, most traces almost entirely overlap). All traces for an algorithm come from a single policy which was trained with $K_{max}=64$ contexts. Where a blank sub-plot appears, the agent was never trained on that context ($K$ was less than $K_{max}$ at the end of 5000 iterations of training).}
\label{xy_point_curriculum}
\end{figure}

\newpage
\subsection{Ant Environment, Uniform Context Distribution, XY-Traces}
\FloatBarrier
\begin{figure}[h]
\centering

\rule{\linewidth}{0.5pt}
VALOR, Uniform Context Distribution: \vspace{2mm}

\begin{subfigure}{0.32\textwidth}
  \includegraphics[width=\textwidth]{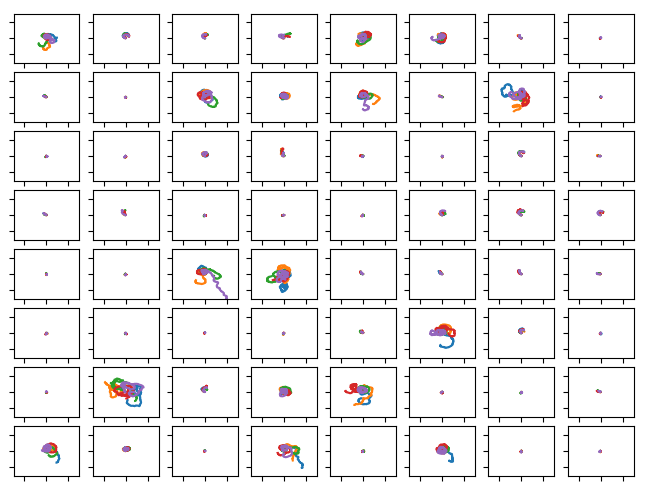}
  \caption{VALOR, Uniform, s0}
\end{subfigure}
\begin{subfigure}{0.32\textwidth}
  \includegraphics[width=\textwidth]{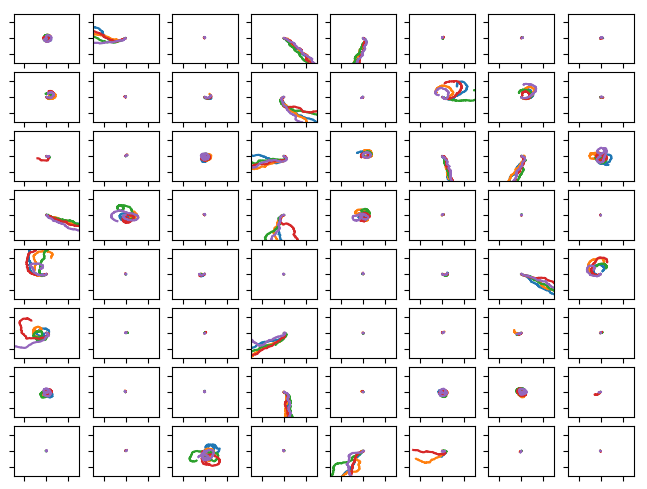}
  \caption{VALOR, Uniform, s10}
\end{subfigure}
\begin{subfigure}{0.32\textwidth}
  \includegraphics[width=\textwidth]{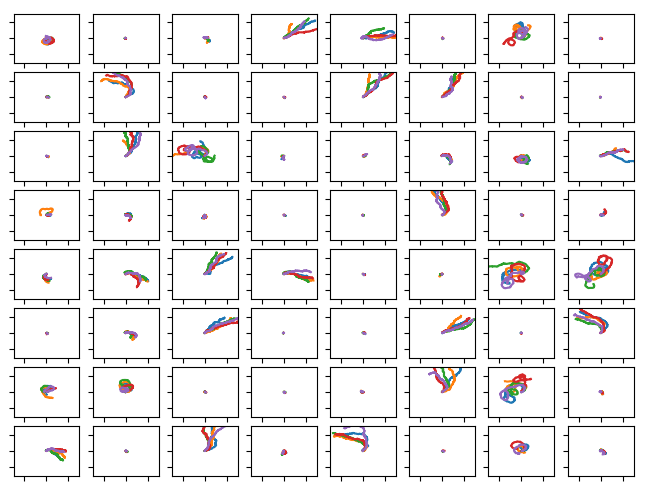}
  \caption{VALOR, Uniform, s20}
\end{subfigure}

\rule{\linewidth}{0.5pt}

VIC, Uniform Context Distribution:\vspace{2mm}

\begin{subfigure}{0.32\textwidth}
  \includegraphics[width=\textwidth]{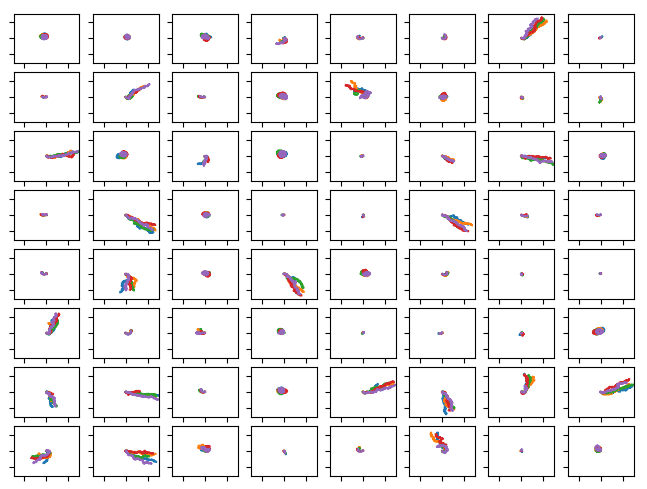}
  \caption{VIC, Uniform, s0}
\end{subfigure}
\begin{subfigure}{0.32\textwidth}
  \includegraphics[width=\textwidth]{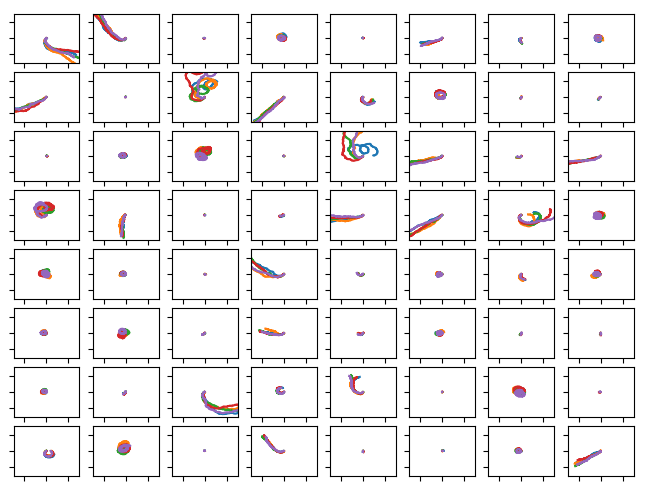}
  \caption{VIC, Uniform, s10}
\end{subfigure}
\begin{subfigure}{0.32\textwidth}
  \includegraphics[width=\textwidth]{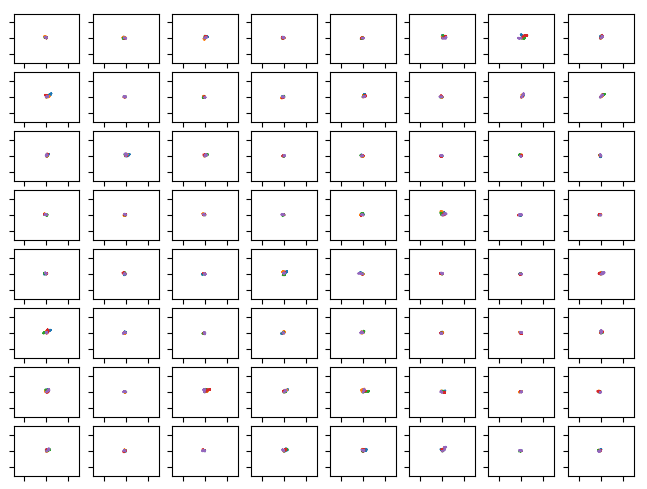}
  \caption{VIC, Uniform, s20}
\end{subfigure}

\rule{\linewidth}{0.5pt}

DIAYN, Uniform Context Distribution:\vspace{2mm}

\begin{subfigure}{0.32\textwidth}
  \includegraphics[width=\textwidth]{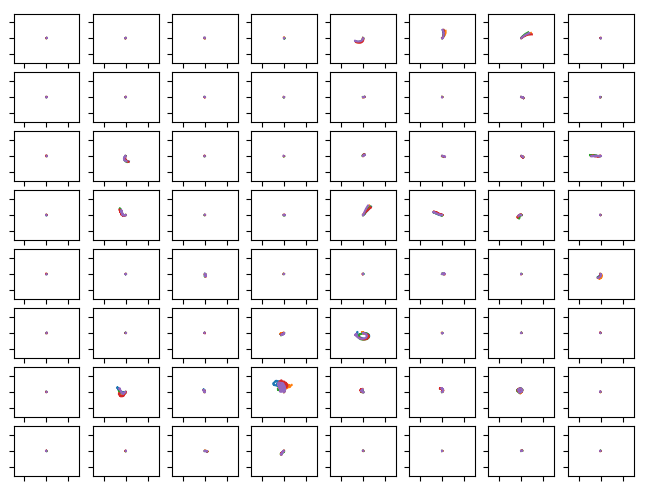}
  \caption{DIAYN, Uniform, s0}
\end{subfigure}
\begin{subfigure}{0.32\textwidth}
  \includegraphics[width=\textwidth]{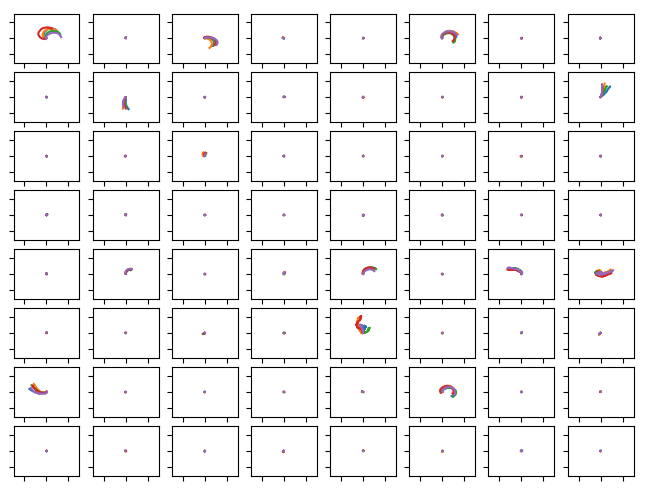}
  \caption{DIAYN, Uniform, s10}
\end{subfigure}
\begin{subfigure}{0.32\textwidth}
  \includegraphics[width=\textwidth]{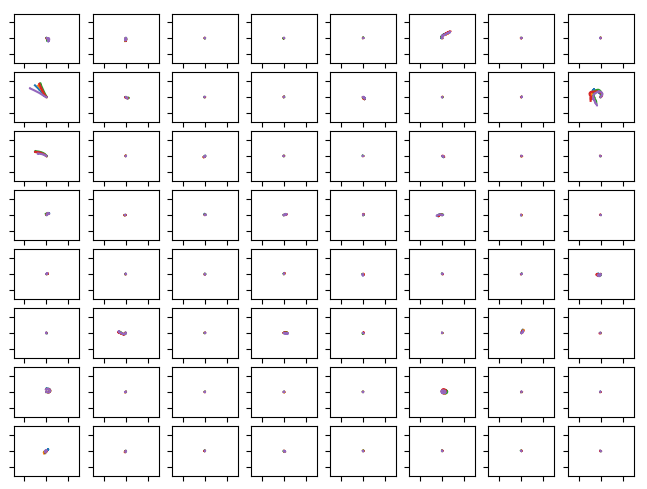}
  \caption{DIAYN, Uniform, s20}
\end{subfigure}

\caption{Learned behaviors in the Ant environment with uniform context distributions. Each sub-plot shows X-Y traces for five trajectories conditioned on the same context (because the learned behaviors are highly repeatable, most traces almost entirely overlap). All traces for an algorithm come from a single policy which was trained with $K=64$ contexts.}
\label{xy_ant_uniform}
\end{figure}

\newpage
\subsection{Ant Environment, Curriculum Context Distribution, XY-Traces}
\FloatBarrier

\begin{figure}[h]
\centering

\rule{\linewidth}{0.5pt}
VALOR, Curriculum Context Distribution: \vspace{2mm}

\begin{subfigure}{0.32\textwidth}
  \includegraphics[width=\textwidth]{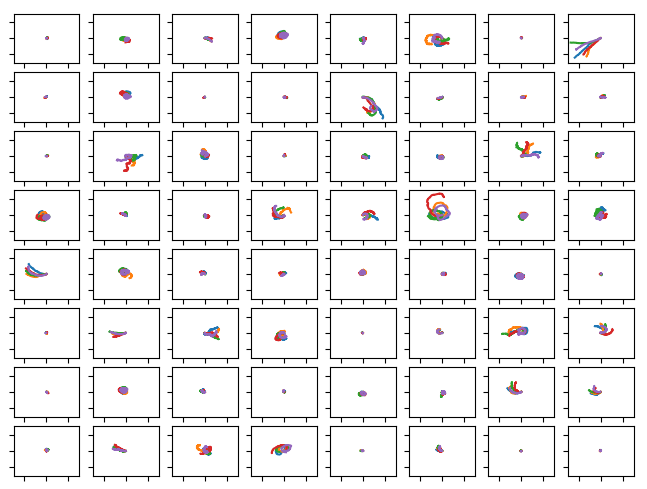}
  \caption{VALOR, Curriculum, s0}
\end{subfigure}
\begin{subfigure}{0.32\textwidth}
  \includegraphics[width=\textwidth]{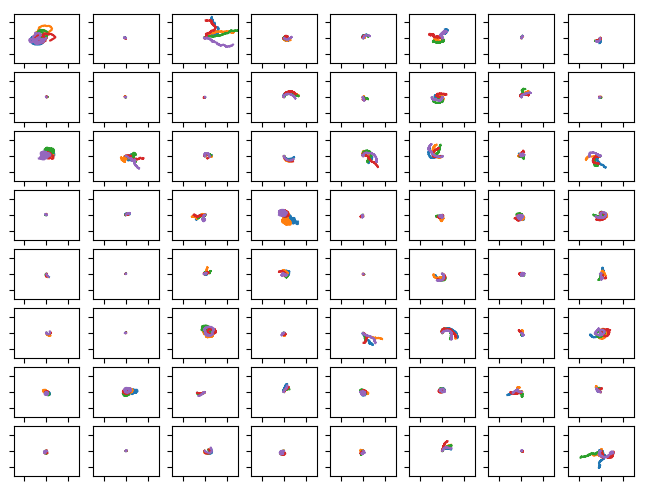}
  \caption{VALOR, Curriculum, s10}
\end{subfigure}
\begin{subfigure}{0.32\textwidth}
  \includegraphics[width=\textwidth]{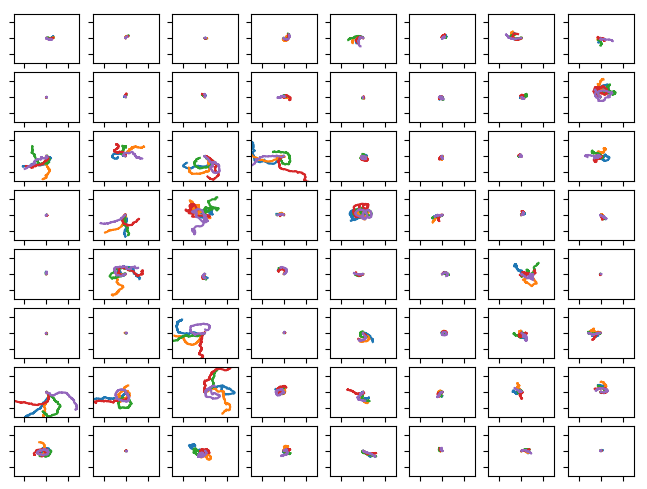}
  \caption{VALOR, Curriculum, s20}
\end{subfigure}

\rule{\linewidth}{0.5pt}

VIC, Curriculum Context Distribution:\vspace{2mm}

\begin{subfigure}{0.32\textwidth}
  \includegraphics[width=\textwidth]{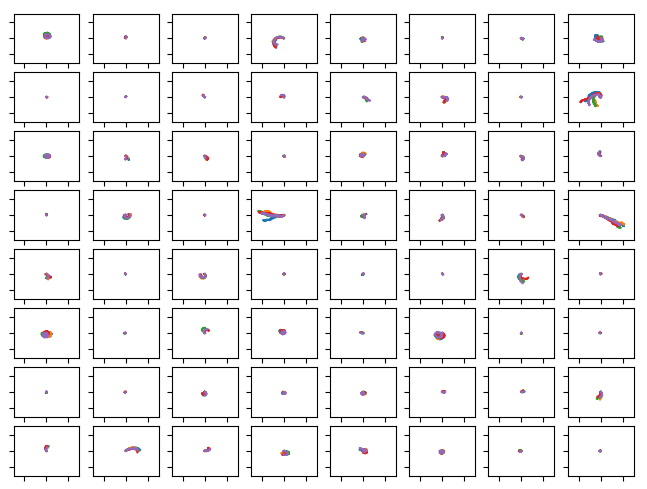}
  \caption{VIC, Curriculum, s0}
\end{subfigure}
\begin{subfigure}{0.32\textwidth}
  \includegraphics[width=\textwidth]{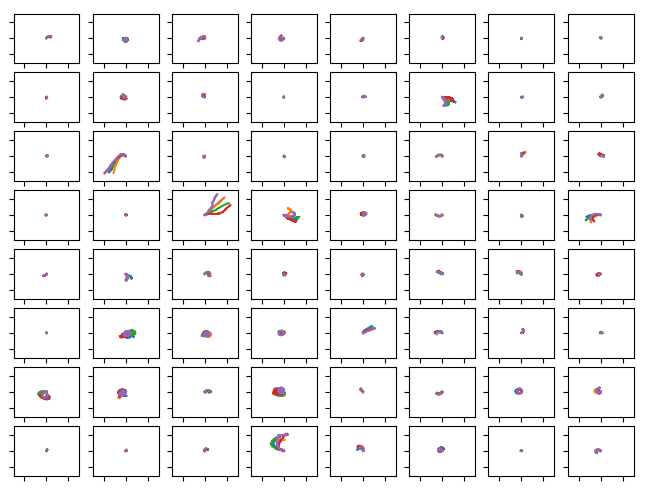}
  \caption{VIC, Curriculum, s10}
\end{subfigure}
\begin{subfigure}{0.32\textwidth}
  \includegraphics[width=\textwidth]{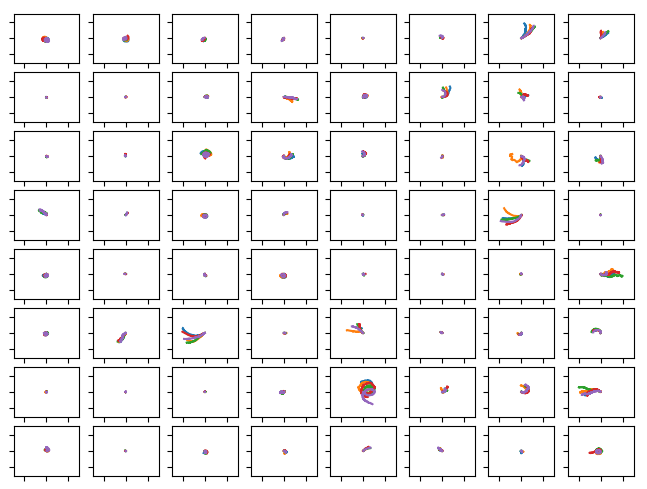}
  \caption{VIC, Curriculum, s20}
\end{subfigure}

\rule{\linewidth}{0.5pt}

DIAYN, Curriculum Context Distribution:\vspace{2mm}

\begin{subfigure}{0.32\textwidth}
  \includegraphics[width=\textwidth]{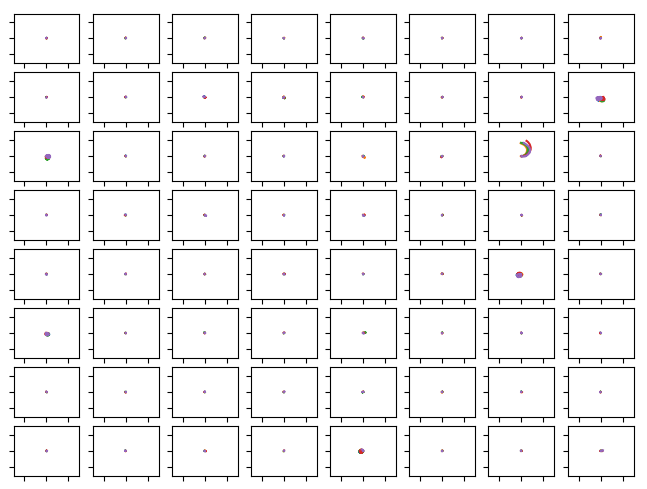}
  \caption{DIAYN, Curriculum, s0}
\end{subfigure}
\begin{subfigure}{0.32\textwidth}
  \includegraphics[width=\textwidth]{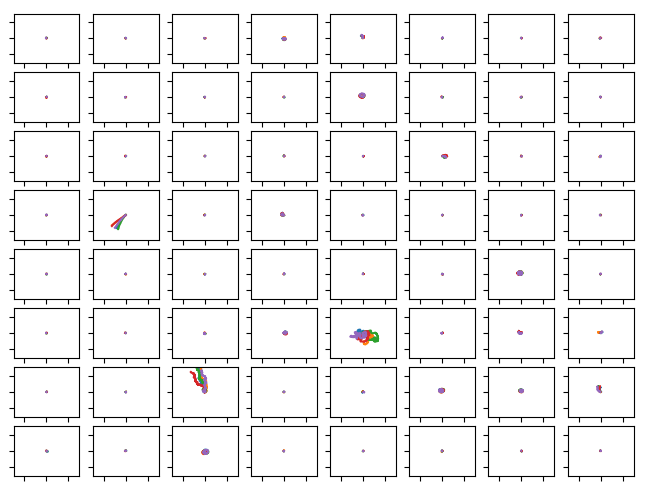}
  \caption{DIAYN, Curriculum, s10}
\end{subfigure}
\begin{subfigure}{0.32\textwidth}
  \includegraphics[width=\textwidth]{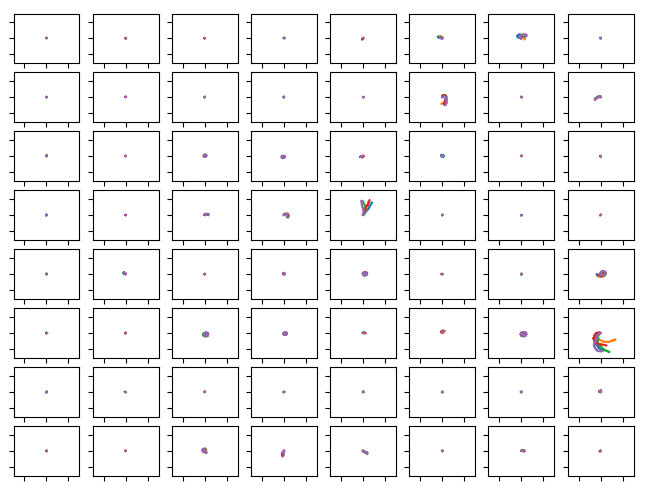}
  \caption{DIAYN, Curriculum, s20}
\end{subfigure}

\caption{Learned behaviors in the Ant environment with the curriculum trick. Each sub-plot shows X-Y traces for five trajectories conditioned on the same context (because the learned behaviors are highly repeatable, most traces almost entirely overlap). All traces for an algorithm come from a single policy which was trained with $K_{max}=64$ contexts. Where a blank sub-plot appears, the agent was never trained on that context ($K$ was less than $K_{max}$ at the end of 5000 iterations of training).}
\label{xy_ant_curriculum}
\end{figure}

\newpage

\section{Learning Multimodal Policies with Random Rewards} \label{random_reward_section}

We considered a random reward baseline, where an agent acting under context $c$ would receive a reward
\begin{equation}
R(s,a,c) = v_c^T s,
\label{random_rewards}
\end{equation}
where $v_c$ was a random context-specific unit vector, obtained by sampling from $\calN(0,I)$ and then normalizing. It seemed plausible that rewards of this form would do a good job of encoding human priors for robot behavior for the simple locomotion tasks in our core comparison. In practice, it turned out to be extremely challenging to train multimodal agents with these rewards; while somewhat easier to train unimodal agents with them, the behaviors that we observed were less interesting than expected. We present results from two sets of experiments:
\begin{enumerate}
\item [RR1.] a ceteris paribus analogue to our core comparison between variational option discovery algorithms, using all of the same hyperparameters (number of epochs, paths per epoch, number of contexts, the use of embeddings, learning rates, etc.), except with rewards from Eq. \ref{random_rewards} instead of a learned decoder,
\item [RR2.] and a set of experiments where all else is equal except that the number of contexts is $K=1$ instead of $K=64$. 
\end{enumerate}

RR1 is a direct and fair comparison, while RR2 allows us to gain intuition for the behavior obtained by optimizing these random rewards separately from the challenges of multitask learning. 

\subsection{Results from RR1}

\begin{figure}[h]
\centering
\includegraphics[width=\linewidth]{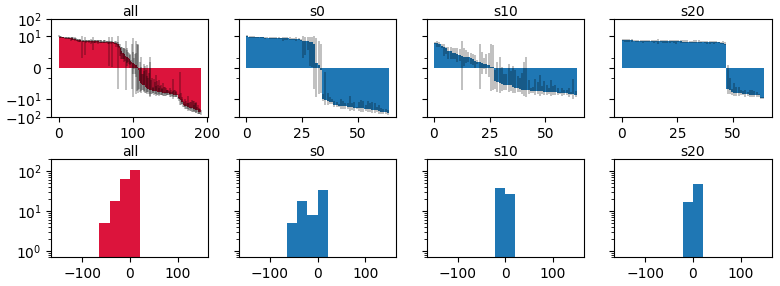}
\caption{Final $x$-coordinate in the Cheetah environment for random rewards.}
\label{random-cheetah}
\end{figure}

\begin{figure}[h]
\centering
\includegraphics[width=\linewidth]{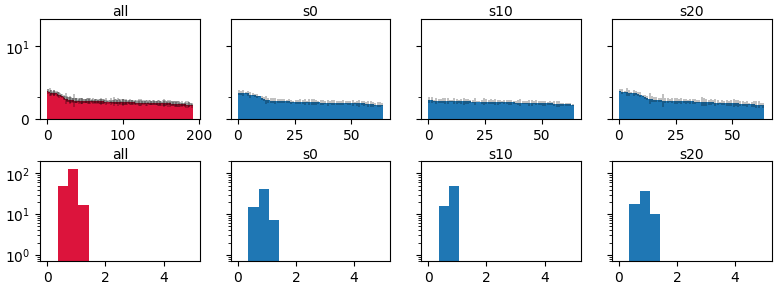}
\caption{Final distance from origin in Swimmer for random rewards.}
\label{random-swimmer}
\end{figure}

\begin{figure}[h]
\centering
\includegraphics[width=\linewidth]{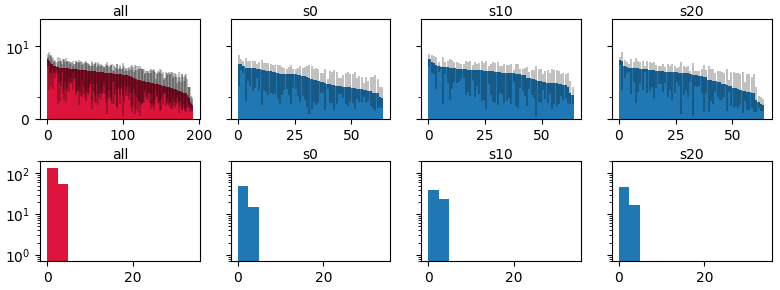}
\caption{Final distance from origin in Ant for random rewards.}
\label{random-ant0}
\end{figure}

\begin{figure}[h]
\centering
\includegraphics[width=\linewidth]{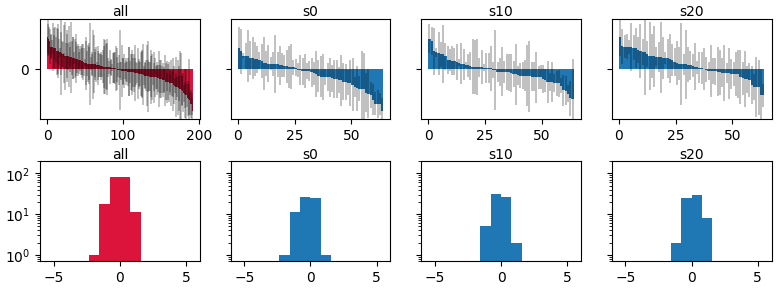}
\caption{Number of $z$-axis rotations in Ant for random rewards.}
\label{random-ant1}
\end{figure}

The results in Cheetah (Fig. \ref{random-cheetah}) look reasonable in composite, but are weak for individual random seeds: in each seed, the results are nearly bimodal, with one mode learning to run forward at some speed, and the other mode learning to run backwards at another speed. In Swimmer (Fig. \ref{random-swimmer}), this form of random rewards inspires almost no motion. Results in the Ant environment (Figs. \ref{random-ant0}, \ref{random-ant1}) show extreme variability: no individual behavior was consistent with respect to the score functions we used (the black bars, representing standard deviation, are very large for every behavior). 

\subsection{Results from RR2}

\begin{figure}[h]
\centering
\includegraphics[width=\linewidth]{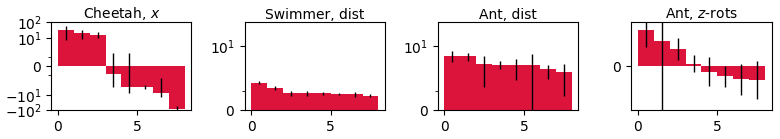}
\caption{Score distributions for RR2.}
\label{rr2}
\end{figure}

We found no significant difference in quality of learned behaviors between the multimodal policies in RR1 and the unimodal policies in RR2, as shown in Fig. \ref{rr2}. That is, training with a single random reward function, instead of several at once, did not result in useful or consistent behavior as measured by our score functions. 

\subsection{Discussion}

Our conclusion is that random rewards based on Eq. \ref{random_rewards} do not result in interesting behavior in the environments we considered. However, there may exist a functional form for random rewards which performs better. 

\end{document}